\documentclass[sigconf, author]{acmart}
\usepackage{multirow}
\usepackage{multicol}
\usepackage{enumitem}
\newcommand{\etal}{{\emph{et al.}}}
\newcommand{\expbf}[1]{+#1$\%$}
\AtBeginDocument{%
  }

\copyrightyear{2023}
\acmYear{2023}
\setcopyright{acmlicensed}
\acmConference[MM '23] {Proceedings of the 31st ACM International Conference on Multimedia}{October 29--November 3, 2023}{Ottawa, ON, Canada.}
\acmBooktitle{Proceedings of the 31st ACM International Conference on Multimedia (MM '23), October 29--November 3, 2023, Ottawa, ON, Canada}
\acmPrice{15.00}
\acmISBN{979-8-4007-0108-5/23/10}
\acmDOI{10.1145/3581783.3612484}

\acmSubmissionID{3441}

\begin{document}

\title{POV: Prompt-Oriented View-Agnostic Learning for \\Egocentric Hand-Object Interaction in the Multi-View World}


\author{Boshen Xu}
\affiliation{%
  \institution{Renmin University of China}
  \country{}
}
\email{xuboshen.uestc@gmail.com}
\author{Sipeng Zheng}
\affiliation{%
  \institution{Renmin University of China}
  \country{}
  }
\email{zhengsipeng@ruc.edu.cn}
\author{Qin Jin}
\authornote{Corresponding Author}
\affiliation{%
  \institution{Renmin University of China}
  \country{}
  }
\email{qjin@ruc.edu.cn}






\renewcommand{\shortauthors}{Boshen Xu, Sipeng Zheng, and Qin Jin}

\newcommand{\qin}[1]{{\color{red}{\bf Qin}: {#1}}}

\begin{abstract}
We humans are good at translating third-person observations of hand-object interactions (HOI) into an egocentric view.
However, current methods struggle to replicate this ability of view adaptation from third-person to first-person.
Although some approaches attempt to learn view-agnostic representation from large-scale video datasets, they ignore the relationships among multiple third-person views. 
To this end, we propose a \textbf{P}rompt-\textbf{O}riented \textbf{V}iew-agnostic learning (POV) framework in this paper, which enables this view adaptation with few egocentric videos. 
Specifically, We introduce interactive masking prompts at the frame level to capture fine-grained action information, and view-aware prompts at the token level to learn view-agnostic representation. 
To verify our method, we establish two benchmarks for transferring from multiple third-person views to the egocentric view. 
Our extensive experiments on these benchmarks demonstrate the efficiency and effectiveness of our POV framework and prompt tuning techniques in terms of view adaptation and view generalization. Our code is available at \url{https://github.com/xuboshen/pov_acmmm2023}.

\end{abstract}





\maketitle
\section{Introduction}
\begin{figure}[h]
	\includegraphics[width=0.38\textwidth]{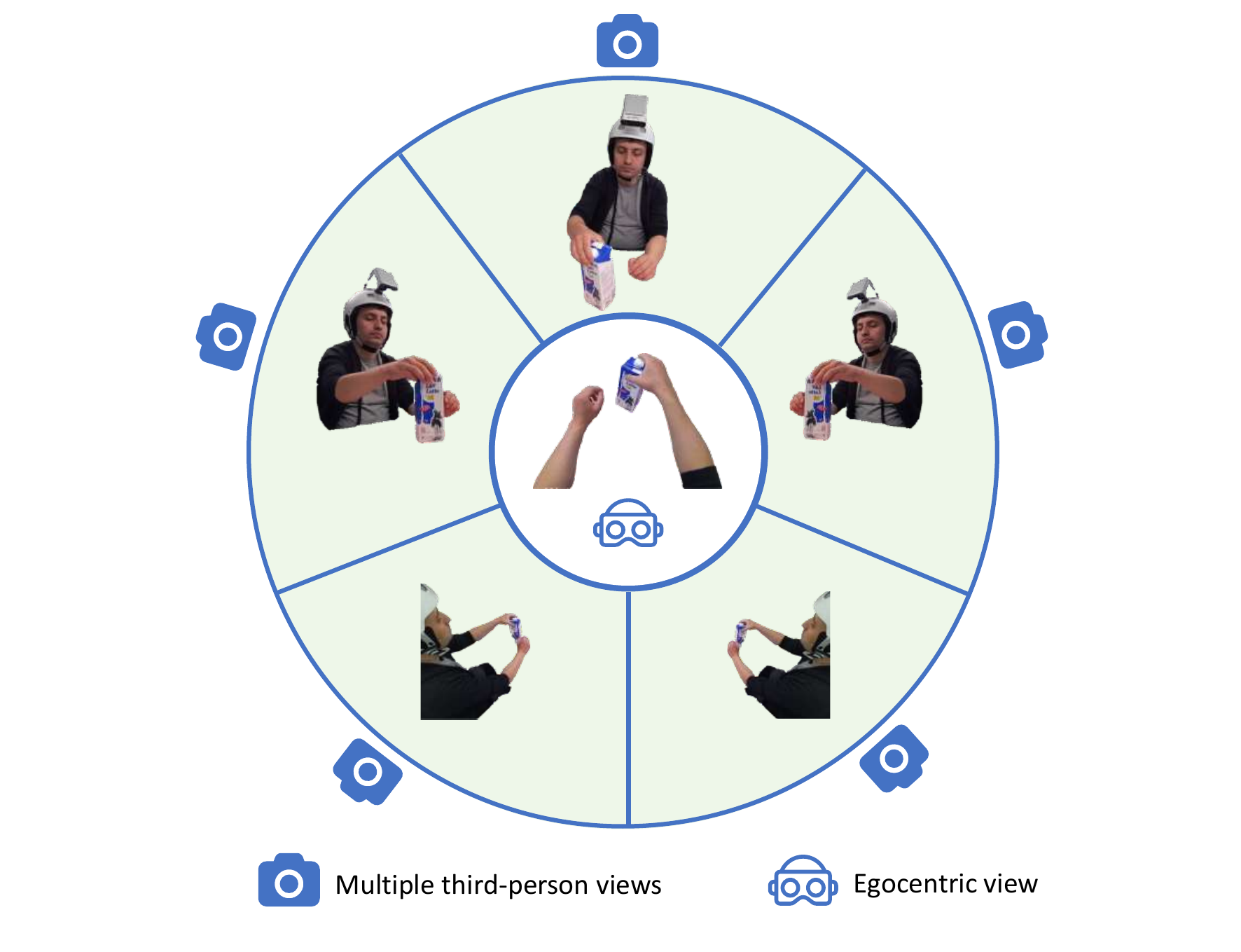}
 \vspace{-8pt}
	\caption{
Humans can learn egocentric hand-object interactions (Ego-HOI) knowledge  by observing extensive third-person videos. Following this intuition, we utilize multi-view third-person videos for learning view-agnostic representation that can be transferred to egocentric view.
 }
	\label{fig:intro}
\end{figure}

In recent years, there has been a growing interest in understanding egocentric hand-object interaction (Ego-HOI)~\cite{yang2022oakink, damen2022rescaling}, which refers to how people interact with objects using their hands from an egocentric view. 
This field has important applications in egocentric robotics vision~\cite{kim2019eyes} and virtual reality~\cite{han2020megatrack}.
Unfortunately, despite the availability of several related benchmarks~\cite{liu2022hoi4d,sener2022assembly101,kwon2021h2o, grauman2022ego4d},
current Ego-HOI works often require bulky laboratory equipment like headset cameras for data collection. 

However, think about us humans, we display an impressive ability to learn how to interact with the real world from a young age by observing the behavior of others.
For example, when we watch a chef preparing a dish on a TV show, it is easy for us to reenact the cooking process ourselves from an egocentric view. 
AI researchers have long striven to empower robots with this same ``do as I show you'' ability, such as enabling an embodied robot to learn hand-object interactions by watching third-person instructional videos on the internet.
In reality, there are rich third-person videos available on the Internet
(e.g., HowTo100M~\cite{miech2019howto100m}).
Therefore, we pose a question: can we learn view-agnostic representations by using a massive amount of third-person data from multiple views to assist in Ego-HOI understanding?

There have been previous endeavors to tackle similar questions. 
To address the issue of the appearance of HOI varying greatly from different third-person views, some works directly employ view-agnostic modalities like 3D pose~\cite{sabater2021domain} and audio~\cite{zhang2022audio}.
However, these approaches have limited scalability and capability due to the high cost of data collection and the lack of visual appearance of objects in HOI tasks. 
Other studies have attempted to learn 3D structures using explicit geometric transformations from RGB inputs~\cite{shang2022learning, piergiovanni2021recognizing, jaderberg2015spatial}. 
However, we argue that these methods require sufficiently abundant views to achieve satisfactory results and struggle to recognize fine-grained interactions.

In this work, we propose a framework called \textbf{P}rompt-\textbf{O}riented \textbf{V}iew-agnostic learning (POV), which develops an effective approach to learning view-agnostic representations from multi-view videos and enabling efficient transfer to the egocentric view.
Our framework incorporates the visual prompt design and prompt tuning strategy into vision transformer\cite{fan2021multiscale}, which are essential to learn view-agnostic representations and avoid overfitting\cite{frankle2018lottery} due to limitations in data scale and diversity. 
Specifically, we introduce \textit{interactive masking prompts} on the frame level, enabling the model to focus on the interaction region from third-person videos, and trainable and lightweight \textit{view-aware prompts} on the token level, empowering the model with view-agnostic knowledge and ability for efficient transfer learning. 
With visual prompts, our model is trained through two essential tasks from third-person videos: prompt-based action understanding and view-agnostic prompt tuning, and one optional task from egocentric videos: egocentric fine-tuning.
The design is partially inspired by how humans learn from observation, which involves first learning from other people's rich instructions (third-person videos) and then exploring egocentric scenarios (egocentric videos) afterward.

To validate our approach, we further define new benchmark settings called Ego-HOI-XView, which utilizes third-person videos during pre-training to help learn HOI knowledge for cross-view fine-tuning and inference in egocentric videos. The benchmarks are based on two multi-view datasets, Assembly101\cite{sener2022assembly101} and H2O\cite{kwon2021h2o}, and are designed to evaluate cross-view egocentric human-object interaction recognition.
We conduct extensive experiments and analyses
on these benchmarks to verify the transferable ability of our model across different views.
Our method outperforms other approaches on all benchmarks.

Our contributions are summarized as follows:
\parskip=0.2em
\begin{itemize}[itemsep=0.2em,parsep=0em,topsep=0em,partopsep=0em,leftmargin=1em,itemindent=0.2em]
\item We propose a view adaption framework called POV that learns view-agnostic representation from third-person videos via learnable and lightweight prompts.
\item We design visual prompts in vision transformers on the frame level and token level to train our model to focus on hand-object interaction and achieve cross-view generalization.
\item We construct two benchmarks on two fine-grained Ego-HOI datasets, Assembly101\cite{sener2022assembly101} and H2O\cite{kwon2021h2o}. 
Our method demonstrates outstanding generalization ability under different downstream Ego-HOI tasks.

\end{itemize}

\section{RELATED WORK}

\noindent \textbf{Egocentric Hand-object Interaction. } 
Egocentric videos~\cite{qiu2021ego, sigurdsson2018charades}, which are usually recorded using head-mounted cameras, offer a distinctive perspective on human activities, especially hand-object interaction (Ego-HOI)~\cite{goyal2017something, damen2022rescaling}.
Despite the extensive research on third-person action recognition in recent years~\cite{ kong2022human, poppe2010survey,herath2017going}, egocentric videos have received less attention.
To improve Ego-HOI recognition, some researchers have developed unique model architectures that utilize 3D CNNs~\cite{lu2019learning}, recurrent networks~\cite{furnari2020rolling, sudhakaran2019lsta}, or multi-stream networks~\cite{kazakos2019epic, li2021eye}.
Despite some progress in developing models for egocentric videos, they still face challenges in handling diverse real-life scenarios because of the high cost of collecting such data.
To address this issue, several datasets have been introduced (e.g., H2O~\cite{kwon2021h2o}, Assembly101~\cite{sener2022assembly101}, and Ego4D~\cite{grauman2022ego4d}), and pre-training methods have been widely discussed (e.g., EgoVLP~\cite{lin2022egocentric}).
However, these datasets have some limitations. 
For example, some lack labels~\cite{grauman2022ego4d}, while others do not account for changes in scenarios~\cite{sener2022assembly101} or are not sufficiently large in scale~\cite{kwon2021h2o}.
To overcome these problems, recent research has focused on leveraging third-person videos to adapt to egocentric views.
Third-person data is more scalable and easier to collect, making it a rich source of information.
For example, Li~\etal~\cite{li2021ego} propose Ego-Exo, which pre-trains the video encoder’s parameters by learning useful visual cues from third-person datasets (e.g., Kinetics~\cite{kay2017kinetics}).
Escorcia~\etal~\cite{escorcia2022sos} introduce SOS, which ensembles an additional structure trained on ImageNet~\cite{deng2009imagenet} to improve the recognition of interacting objects in HOI.
In addition, Choi~\etal~\cite{Choi_2020_WACV} propose an unsupervised domain adaptation method for Ego-HOI by learning from labeled third-person videos and unlabeled egocentric ones.
Building on this line of research, our work focuses on using extra third-person videos to achieve view adaptation for Ego-HOI.

\begin{figure*}[ht]	
 \includegraphics[width=1\textwidth]{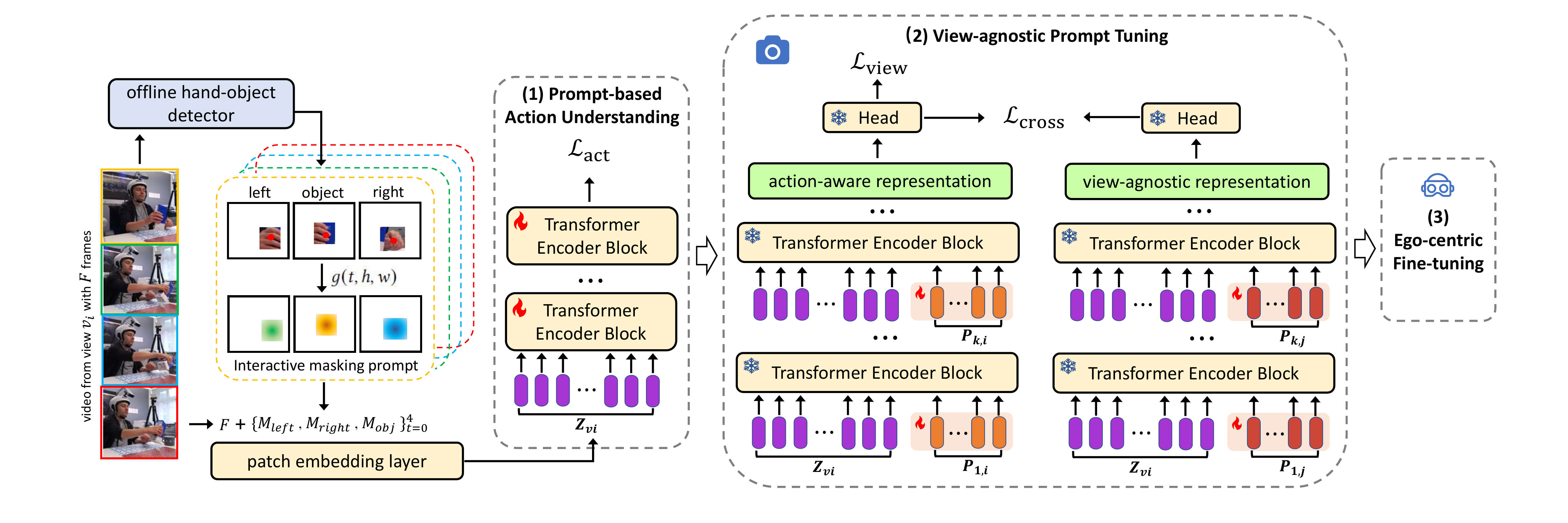}
  \vspace{-8pt}
\caption{Illustration of our prompt-oriented view-agnostic learning framework, which trains a model through two optimization sub-tasks and one optional sub-task: 
(1) prompt-based action understanding, which incorporates interactive masking prompts into frames to pre-train the entire model on third-person videos;
(2) view-agnostic prompt tuning, where only view-aware prompts are fine-tuned through cross-view alignment and cross-entropy loss. 
(3) egocentric fine-tuning, where the model is optionally fine-tuned on limited egocentric videos. 
}
	\label{fig:arch}
\end{figure*}

\noindent \textbf{View-agnostic Representation Learning. } 
Learning view-agnostic representation is essential to adapting third-person data to egocentric scenarios, which has been a widely discussed challenge~\cite{rao2001view, farhadi2008learning, shen2009view, ji2019attention} for a long time.
This concept has various applications in downstream tasks, including pose estimation~\cite{sun2020view}, robotics~\cite{jangir2022look, sharma2019third}, and 3D detection~\cite{li2022bevformer, liu2022petr}.
Some early approaches~\cite{sigurdsson2018actor, shah2023multi, ji2019attention} utilize synchronized multi-view data to address the challenge of view-agnostic representation.
However, the high cost of such data makes these approaches barely scalable for broader applications.
Instead, recent studies have explored the relationship between different views by utilizing unsynchronized multi-view data, which is much easier to collect.
Some of these studies~\cite{piergiovanni2021recognizing, shang2022learning, jaderberg2015spatial} investigate the learning of 3D structures from images by applying explicit geometric transformations in latent space within CNNs~\cite{carreira2017quo} or Transformers~\cite{arnab2021vivit}.
Furthermore, some works propose generating unseen latent view representations from third-person videos that have already been seen.
For example, Das ~\etal~\cite{das2023viewclr} propose ViewCLR to maximize the similarities
between the representation from original viewpoint with its latent unseen viewpoint representation.
Similarly, Vyas~\etal\cite{vyas2020multi} utilize auxiliary tasks of cross-view prediction to learn view-agnostic knowledge by rendering frames from unseen views.
The previous works mainly focus on human body activities between third-person views, while in this work, we explore the relationship between unsynchronized multi-view videos towards Ego-HOI.

\noindent \textbf{Visual Prompt Tuning.}
In academic literature, prompts~\cite{liu2023pre} are typically defined as hand-crafted text instructions~\cite{shin2020autoprompt, jiang2020can} that are placed at the beginning of input text. 
These prompts help large language models to better comprehend the task at hand.
However, recent research~\cite{lester2021power, li2021prefix} has begun treating prompts as task-specific continuous vectors.
These vectors can be optimized through gradient descent during the fine-tuning process, which is known as prompt tuning.
Additionally, the prompt-based learning approach has demonstrated promising outcomes in various vision tasks.
For the first time, Jia~\etal~\cite{jia2022visual} introduce VPT, a method that efficiently adapts pre-trained models for image classification by concatenating trainable prompt tokens with image tokens in vision transformers. Similarly, Zhang~\etal~\cite{zhang2022can} propose DepthCLIP, which fine-tunes a vision-language model with prompts to generate visual concepts for monocular depth estimation. 
To address the issue of distribution shift caused by out-of-domain data, 
Gao~\etal~\cite{gao2022visual} propose DePT, which explores self-supervised prompt tuning.
Zheng~\etal~\cite{zheng2022prompt} propose DoPrompt, which embeds domain-specific knowledge into prompts to predict out-of-domain data.
Moreover, ProTrack~\cite{yang2022prompting} and ViPT~\cite{zhu2023visual} introduce modality-specific prompts for applying pre-trained models to downstream tracking tasks.
In this study, the emphasis is on utilizing prompt tuning to develop a view-agnostic representation for Ego-HOI.


\section{Method Overview}
Collecting data from an egocentric perspective can be expensive, so using rich third-person videos to aid understanding is a practical alternative.
This paper focuses on the egocentric hand-object interaction (Ego-HOI) task, specifically under the \textbf{\textcolor{blue}{Cross-View}} setting, which we call the \textbf{Ego-HOI-XView} task.
In this task, the model is first trained on third-person videos $D_{\rm 3rd}$ (\textcolor{blue}{\textbf{Source View}}) from various camera angles in $\mathcal{V}_{\rm 3rd}=\{v_1, v_2,...,v_N\}$, then optionally fine-tuned on egocentric scenarios $\mathcal{V}_{\rm ego}$ (\textcolor{blue}{\textbf{Target View}}) with unlabeled (zero-shot) or limited labeled data (few-shot) $\hat{D}_{\rm ego}$, and finally used to make predictions in the target view.
We believe that this task setting closely mimics how humans learn:
start by learning from other people's rich instructions (represented by $D_{\rm 3rd}$), then explore and experiment on our own (represented by $\hat{D}_{\rm ego}$), and finally, gain the ability to understand different skills from an egocentric view with minimal hands-on instruction in reality.

In this paper, we propose a framework called \textbf{P}rompt-\textbf{O}riented \textbf{V}iew-agnostic learning (\textbf{POV}) to address the Ego-HOI-XView task.
Our main argument is that to achieve success in the Ego-HOI-XView task, it is essential to develop view-agnostic knowledge for video data.
This means that the model should be able to understand human-object interactions (HOIs) from any camera angle, particularly egocentric ones. 
In Section~\ref{sec:framework}, we describe how POV can be used to realize this goal.
Moreover, we observe that the view-agnostic tuning objective is challenging to achieve due to limited data in terms of scale and diversity, leading to inherent noise in the objective.
Therefore, training this objective with full parameters is prone to overfitting.
Bearing this in mind, our POV framework uses visual prompt tuning to adapt the model to different camera angles with minimal parameters.
To further clarify the training and inference procedures for pre-training and downstream tasks, we provide detailed explanations in Section~\ref{inference}.

\section{POV Framework}
\label{sec:framework}
In a typical HOI task, a model analyzes $T$ frames 
$F$ extracted from a third-person video in $D_{\rm 3rd}$ to predict $y=\{y_v, y_o\}$. 
Here, 
$F\in \mathbb{R}^{T\times H \times W \times 3}$ refer a set of frames extracted from the video.
Meanwhile, $y_v$, $y_o$ and $y$ refer to the categories of verb, object, and verb-object combination, respectively. 
In our POV framework, as shown in Figure~\ref{fig:arch}, we utilize a vision transformer model $\mathcal{G}$ with parameters $\Theta$.
Training POV involves two essential tasks and one optional task.
We pre-train our model on third-person videos using two essential learning tasks: prompt-based action understanding and view-agnostic prompt tuning. 
Then, we optionally fine-tune the model for egocentric scenarios through egocentric fine-tuning.

\subsection{Prompt-based Action Understanding} 
It is important to note that action-related knowledge, especially for hand-object interactions, is highly detailed and fine-grained.
For instance, the actions "grab milk" and "squeeze lotion" belong to two distinct HOI categories, even though they have similar visual appearances with only subtle differences.
Directly learning subtle representations is challenging, requiring large-scale high-quality data with labeling that captures the subtle differences.
As a result, we consider training POV to learn fine-grained HOI via prompts. 
Below, we first introduce our proposed \textit{interactive masking prompt} and then present the learning objective with this prompt.

\noindent\textbf{Interactive Masking Prompt. }
As illustrated on the left side in Figure~\ref{fig:arch}, the interactive masking prompts are added to video frame inputs at the frame level to highlight important regions of hand-object interactions.
Each interactive masking prompt $M_i\in \mathbb{R}^{T\times H\times W}$, where $i\in\{{\rm left}, {\rm right}, {\rm object}\}$, has the same scale as the input frames.
To generate the prompts during POV pre-training on $D_{\rm 3rd}$, we use an off-the-shelf detector~\cite{shan2020understanding} to localize the center point $c_i(x_{c_i}, y_{c_i})$ with the same bounding box side $|c|$ of the left hand, right hand and interacting object respectively at each frame.
Then each interactive masking prompt can be generated as:
\begin{equation}
\left[M_{i}\right]_{t, h, w}=\left\{\begin{array}{l}
g(t, h, w), \text { if }\left|h-x_{c_{i}}\right| \leq|c| \text , \left|w-y_{c_{i}}\right| \leq|c| \\
0, \text { otherwise }
\end{array}\right.
\end{equation}
The value of the pixel at $(t,h,w)$ within the important region is determined by the function $g(\cdot)$, and set to zero if it is not within the region.
We introduce two types of interactive masking prompts: hard prompts and soft prompts.
The hard prompt, similar to Jaderberg~\etal~\cite{jaderberg2015spatial} but without geometric transformation, acts as a frame-level indicator to enable the model to capture the interaction region. 
For a hard prompt, each value in the important region is fixed and initialized by a Gaussian distribution: $g(t,h,w) = n\sim\mathcal{N}(0, 1)$.
In contrast, for a soft prompt, each value is a learnable parameter: $g(t,h,w) = \theta$.
Note that the interactive masking prompts are added only during pre-training, thus it is not required during inference.

\begin{table*}
\caption{Comparison results of zero-shot Ego-HOI-XView recognition, where ``3rd'' represents only fine-tuning pre-trained MViT (e.g., Kinetics) on Assembly101 and H2O with Cross-Entropy loss.}
\vspace{-8pt}
\scalebox{0.9}{
\begin{tabular}{c|ccccccc@{}|cccccc}
\toprule
&\multicolumn{6}{c}{Assembly101}&
& 
\multicolumn{6}{c}{H2O}\\
\midrule
\multirow{2}{*}{Method} & \multicolumn{2}{c}{verb} & \multicolumn{2}{c}{noun} & \multicolumn{2}{c}{action}&&
\multicolumn{2}{c}{verb} & \multicolumn{2}{c}{noun} & \multicolumn{2}{c}{action}\\
& top-1 & top-5 & top-1 & top-5 & top-1 & top-5&&
top-1 & top-5 & top-1 & top-5 & top-1 & top-5\\
\midrule
MViT-3rd & 17.00 & 45.27 & 14.42 & 36.61 & 4.64 & 17.33 &&40.62 & 83.39 & 77.73 & 93.16 & 37.89 & 78.51\\
SHOT~\cite{liang2020we} & 26.15 & 75.63 & 11.86 & 46.71 & 5.56 & 18.90 &&
32.61 & 75.78 & 73.43 & 94.33 & 27.73 & 69.72\\
AaD~\cite{yang2022attracting} & 20.66 & 75.48 & 12.57 & 37.55 & 3.75 & 14.20 &&
13.50 & 49.90 & 12.32 & 53.22 & 3.32 & 16.24\\
Tent~\cite{wang2021tent} & 22.70 & 76.91 & 10.78 & \textbf{55.55} & 2.17 & 23.87 &&
48.24 & 85.54 & 75.97 & 93.94 & 43.16 & 79.68 \\
\midrule
POV & \textbf{29.06} & \textbf{80.23} & \textbf{18.23} & 50.73 & \textbf{8.21} & \textbf{27.90} &&
\textbf{50.97} & \textbf{88.08} & \textbf{79.68} & \textbf{95.50} & \textbf{46.67} & \textbf{83.00}\\
\bottomrule
\end{tabular}}
\label{tab:zeroshot}
\end{table*}

\noindent\textbf{Prompt-based Action Understanding Loss.}
We pre-train all parameters $\Theta$ of POV via a sub-task that aims to learn fine-grained Ego-HOI knowledge on third-person videos using a supervised objective like Cross-Entropy loss.
To achieve the goal, POV must focus on the essential regions, such as the hands and the interacting object, while filtering out background noise.
To capture such region-level visual cues, we leverage our interactive masking prompts: $\{M_{\rm left}, M_{\rm right}, M_{\rm obj}\}$.
These prompts are added to the video frames $F$, which serve as inputs to the patch embedding of vision transformers.
The prompt-based action understanding learning objective can be written as follows, where $\delta(\cdot)$ denotes the SoftMax function:
\begin{equation}
    \mathcal{L}_{\rm act} = -y \log(\delta(\mathcal{G}(F, M_{\rm left}, M_{\rm right}, M_{\rm obj}; \Theta))))
\end{equation}

\subsection{View-agnostic Prompt Tuning}
After training with the action understanding objective,
our model gains the ability to extract fine-grained HOI knowledge from third-person videos.
Since the way a human appears to interact with an object can differ greatly from various views, it is essential to ensure that the extracted action-aware knowledge is view-invariant. Therefore, in this phase, we aim to train POV to learn view-agnostic representation from multi-view videos via \textit{view-aware prompts}. 
Please note that the view labels, if not available, can be identified inexpensively using an off-the-shelf view classifier~\cite{lingtao2019object}.
Meanwhile, there exist multiple third-person video corpora that contain different camera view angles, such as Kinetics~\cite{kay2017kinetics}, TSU~\cite{das2019toyota}, and NTU\cite{liu2020ntu}, which can all be leveraged for learning view-agnostic knowledge. 
\ \newline
\noindent\textbf{View-aware prompt. } 
We introduce trainable view-aware prompts $P$ that are added to video tokens.
We divide the transformer layers into $B$ blocks, where each block corresponds to a group of continuous transformer layers.
We denote $\mathcal{B}$ as the set of the first transformer layer IDs of each block.
Specifically, considering a first transformer layer $b_j\in\mathcal{B}$ in a block, a group of view-aware prompts $P_{i, j}=\{p_{i, j}^1,p_{i, j}^2,\cdots,p^n_{i, j}\}$, representing the view information $v_i$ of video $V$, is concatenated with the video tokens from the video $V$ as input throughout the following block.
Let the visual model $\mathcal{G}$ be the composition of structures in vision transformer: $\mathcal{H}\circ \mathcal{T}_B\circ \dots \circ \mathcal{T}_1\circ \mathcal{E}$, where $\mathcal{H}$, $\mathcal{T}_j$ and $\mathcal{E}$ refer to the patch embedding layer, the $j$-th transformer block, and the classification head, respectively. 
Given video $V$ with view $v_i$, the video frames are firstly passed to the patch embedding layer $\mathcal{E}$ and the output is $Z_1 = \mathcal{E}(F)$. 
Then, the prompt augmented vision transformer with $B$ blocks is as follows:

\begin{equation}
    \begin{split}
        [Z_{j+1}, z_{j+1}] &= \mathcal{T}_j([Z_{j}, P_{i, j}]),\ \ j=1,2,\cdots, B\\
        \hat{y} &= \mathcal{H}(Z_{l+1})\\
    \end{split}\label{prompt_vit}
\end{equation} 

where $Z_j$ represents the video tokens input, $z_{j}$ represents the prompt token input in the $j$-th transformer block, and $[\cdot]$ denotes the concatenation operation respectively. 
Moreover, we define model-level prompts and layer-level prompts, which are also called shallow prompts and deep prompts in Jia~\etal~\cite{DBLP:conf/eccv/VPT22}, as a special case of block-level prompts where the $\mathcal{B}=\{1 \}$ and $\mathcal{B}=\{1,2,\dots l \}$.

\noindent\textbf{View-agnostic Prompt Tuning Loss.}
To retain the discriminability of the original video embeddings with viewpoint label $v_i$, we adopt Cross-Entropy loss, which is written as:
\begin{equation}
    \mathcal{L}_{\rm view} =  -y\log(\delta(\mathcal{G}(F_{v_i}, P_i; \Theta_{p}))) 
\end{equation}
where $\Theta_{p}$ represents that all parameters except prompts are frozen during view-agnostic tuning to mostly retain the action knowledge and $P_i$ denotes the collection of $\{P_{i,1}, P_{i, 2}, \cdots P_{i, B}\}$.  
The view-aware prompts are added on the token level as described in equation~\ref{prompt_vit}, as denoted here for the sake of clarity and simplicity.

During view-agnostic prompt tuning, for video frames $F_{v_i}$ with a camera view angle of $v_i$, we utilize a cross-view alignment loss to minimize the video embedding distance between its original view $v_i$ and an unseen view $v_j$, where the view is simulated by view-aware prompts. 
The criterion can be formulated as:

\begin{equation}
    \mathcal{L}_{\rm cross}(i,j) = D_{KL}(\mathcal{G}(F_{v_i}, P_i; \Theta_{p}), \mathcal{G}(F_{v_i}, P_j; \Theta_{p}))
\end{equation}

where $D_{KL}$ denotes the KL-divergence distance.
The final cross-view alignment loss for a given video is written as:

\begin{equation}
    \mathcal{L}_{\rm cross} = \sum_{v_j\in \mathcal{V}_{\rm 3rd}, j\ne i}\mathcal{L}_{\rm cross}(i, j)
\end{equation}

With the aid of visual prompt tuning, our model is less prone to overfitting and can learn view-agnostic representations in the feature space for Ego-HOI recognition.

\subsection{Egocentric Fine-Tuning}
After being pre-trained on $D_{\rm 3rd}$, POV enhances the model's ability to adapt to egocentric scenarios in the downstream Ego-HOI-XView task, using either unlabeled or limited labeled egocentric data $\hat{D}_{\rm ego}$.

\noindent\textbf{Joint-view prompt.}
In the fine-tuning and final inference stages on novel views such as the egocentric view $\hat{D}_{\rm ego}$, we first ensemble the view-aware prompts at each block.
This generates a joint prompt $P_{b}=\sum_{v_i\in \mathcal{V}_{\rm 3rd}} P_{i,b}$ that combines the view-aware prompts from each view  $V_i$ in $\mathcal{V}_{\rm 3rd}$, and enables efficient leveraging of knowledge from the view-aware prompts.
Next, we feed the joint prompts and video tokens from the novel views $v^{*}$, such as the egocentric view or unseen third-person view, to the transformer layers.
The input sequence, denoted as $[Z_{j, v^{*}}, P_{b,j}],j\in [1,2,\cdots]$, is then passed through the vision transformer, following Equation~\ref{prompt_vit}.

\noindent\textbf{Egocentric Fine-tuning Loss.}
The model can be fine-tuned either with unlabeled data using prompt tuning or with limited labeled data using full tuning.
For prompt tuning, the optimization process can use any feasible self-supervised objective such as information maximization introduced in \cite{liang2020we}:
\begin{equation}
    \mathcal{L}_{\rm ego} = \mathcal{L}_{\rm ssl}(\mathcal{G}(F_{\rm ego}, P_{b}; \Theta_p))
\end{equation}

For full tuning, we use a standard Cross-Entropy loss to improve action understanding on egocentric videos in a supervised manner. 
\begin{equation}
    \mathcal{L}_{\rm ego} = -ylog(\delta(\mathcal{G}(F_{\rm ego}, P_b; \Theta)))
\end{equation}
where $F_{\rm ego}$ denotes egocentric video frames.
For simplicity, we use $\Theta_{\rm ego}$ to represent the optimized parameters that are adapted to the type of fine-tuning used, namely prompt tuning or full tuning.

\subsection{Training Strategy} 
\label{inference}
POV pre-trains our model on $D_{\rm 3rd}$ by first carrying out prompt-based action understanding and then view-agnostic prompt tuning.
In the first learning task, we optimize the whole parameters of the model by using labeled third-person videos, which is achieved through the following objective:
\begin{equation}
\Theta^* = \mathop{argmin}\limits_{\Theta} \mathcal{L}_{\rm act}
\end{equation}

In view-agnostic prompt tuning, we keep the pre-trained parameters $\Theta^*$ frozen.
Only the parameters of newly added prompts are updated with the following objective:
\begin{equation}
\Theta_p^* = \mathop{argmin}\limits_{\Theta_p} (\mathcal{L}_{\rm view} + \lambda \mathcal{L}_{\rm cross})
\end{equation}

The hyper-parameter $\lambda$ controls the weight of the view alignment loss. 
To ensure stable prompt tuning, we set a small value (e.g., 0.001) for $\lambda$.
Furthermore, in the optional fine-tuning task on the egocentric view, the objective is as follows:

\begin{equation}
    \Theta_{\rm ego}^* = \mathop{argmin}\limits_{\Theta_{\rm ego}} \mathcal{L}_{\rm ego}
\end{equation}

\begin{table}[t]

\caption{Comparison results of few-shot Ego-HOI-XView recognition. We report top-1 accuracy here.}
\vspace{-8pt}
\scalebox{0.9}{
\begin{tabular}{c|cccc@{}|ccc}
\toprule
&\multicolumn{3}{c}{Assembly101} &&
\multicolumn{3}{c}{H2O}\\
\midrule
\multirow{1}{*}{Method} & 
verb & object & action &&verb & object &action \\
\midrule
EgoVLP~\cite{lin2022egocentric}  & 5.71 & 3.98 & 0.53 &&
34.63 & 27.59 & 13.30\\
Ego-exo~\cite{li2021ego} & 47.92 & 44.68 & 29.09 &&
46.18 & 59.88 & 35.22\\  
\midrule
POV & \textbf{58.76} & \textbf{54.89} & \textbf{46.06} &&
\textbf{77.53} & \textbf{75.78} & \textbf{73.04}\\
\bottomrule
\end{tabular}}

\label{tab: few_shot}
\end{table}
\begin{table*}
\caption{Ablation comparison of 3rd-to-ego HOI recognition.}
\vspace{-8pt}
\scalebox{0.9}{
\begin{tabular}{c|ccccccc@{}|cccccc}
\toprule
&\multicolumn{6}{c}{Assembly101} && \multicolumn{6}{c}{H2O}\\
\midrule
\multirow{2}{*}{Method} & \multicolumn{2}{c}{verb} & \multicolumn{2}{c}{noun} & \multicolumn{2}{c}{action}&&
\multicolumn{2}{c}{verb} & \multicolumn{2}{c}{noun} & \multicolumn{2}{c}{action}\\
& top-1 & top-5 & top-1 & top-5 & top-1 & top-5&& top-1 & top-5 & top-1 & top-5 & top-1 & top-5\\
\midrule
MViT-frozen & 12.65 & 41.89 & 5.74 & 25.91 & 0.91 & 4.68&
& 16.99& 41.79 & 9.76 & 42.18 & 2.73 & 15.03\\
MViT-3rd & 17.00 & 45.27 & 14.42 & 36.61 & 4.64 & 17.33&
& 40.62 & 83.39 & \textbf{77.73} & 93.16 & 37.89 & 78.51 \\
EgoVLP~\cite{lin2022egocentric} & 11.61 & 40.24 & 6.59 & 15.46 & 0.94 & 4.48 &&
40.50 & 81.23 & 27.78 & 56.16 & 13.50 & 42.27\\
3DTRL~\cite{shang2022learning} & 20.46 & 49.82 & 11.94 & 39.03 & 4.08 & 15.64&&
38.47 & 81.25 & 36.71 & 78.90 & 27.53 & 64.84 \\
\midrule
POV w/o ego & \textbf{30.33} & \textbf{77.06} & \textbf{15.41} & \textbf{51.22} & \textbf{6.78} & \textbf{23.73}&&
\textbf{48.24} & \textbf{85.93} & 77.53 & \textbf{94.72} & \textbf{43.35} & \textbf{79.88}  \\ \bottomrule
\end{tabular}}
\label{tab:abl_2_3rd-to-ego}
\end{table*}


\section{Experiments}

\subsection{Experiment Setup}
\noindent\textbf{Datasets. }
We use two datasets, Assembly101~\cite{sener2022assembly101} and H2O~\cite{kwon2021h2o}, to evaluate our POV framework.
Assembly101 consists of over one million video snippets that show people assembling and disassembling toys, which involves 1,380 HOI categories in total. We select 142 categories in this work.
For training and testing, we select samples from the original training and validation sets, respectively.
To pre-train our model, we use a subset of 134,104 third-person video snippets as $D_{\rm 3rd}$ from Assembly101.
We also use a subset of 10,388 egocentric snippets for egocentric fine-tuning and another 10,388 snippets for testing.
Additionally, we select 20,766 snippets from two third-person camera views that are not seen during training for the HOI-XView evaluation.
Regarding the H2O dataset, which has 36 fine-grained HOI categories, we sample 2,764 third-person video snippets for third-person pre-training, 180 and 511 egocentric snippets for fine-tuning and testing.

\noindent\textbf{Evaluation Metrics. } 
We follow the previous evaluation protocol ~\cite{damen2018scaling} and measure the prediction accuracy for the verb, object, and action respectively. 
The action is defined as the combination of the verb and object.
We report both top-1 and top-5 accuracy.
We test the model's ability to adapt to different views under either zero-shot~\cite{liang2020we} or few-shot setups.

\noindent\textbf{Implementation Details. }
We initialize our POV using an MViT-S model~\cite{fan2021multiscale} that has been trained on Kinetics~\cite{carreira2017quo}. 
For each video, we randomly select four frames as input and resize them to 224 $\times$ 224. 
In addition, we use various data augmentation techniques such as random horizontal flipping and multi-crop~\cite{wang2016temporal}.
During pre-training on $D_{\rm 3rd}$, we employ the AdamW optimizer and cosine scheduler.
The learning rate is initially set to 1e-3 and decreases to 1e-5 by the end of training.
We use the hard type of interactive masking prompt with a size of $1\times 1$ for Assembly101 and $20\times 20$ for H2O.
We incorporate block-level view-aware prompts with $\mathcal{B}=\{1, 2, 4, 15\}$ and prompt number $n=2$ for both datasets.

\begin{table}[t]
\caption{Ablation comparison of zero-shot HOI-XView recognition, where ``frozen'' represents the model pre-trained on Kinetics and without any fine-tuning, POV w/o ego represents POV without egocentric fine-tuning. We report top-5 accuracy on Assembly101.}
\vspace{-8pt}
\scalebox{0.9}{
\begin{tabular}{c|ccc}
\toprule
\multirow{1}{*}{Method} & 
verb & object & action \\
\midrule
MViT-frozen & 68.16 & 45.11  & 5.07 \\
MViT-3rd & 90.79 & 84.79 & 66.95 \\
3DTRL~\cite{shang2022learning} & 90.45 & 84.57 & 67.80 \\
\midrule
POV w/o ego & \textbf{91.19} & \textbf{85.03} & \textbf{69.09} \\
\bottomrule
\end{tabular}
}
\label{tab:abl_3rd}
\end{table}

\subsection{Comparison with State of the Arts}

\noindent\textbf{Zero-shot Ego-HOI-XView Recognition. }
The zero-shot setup involves using unlabeled egocentric videos of $\mathcal{V}_{\rm ego}$ for egocentric fine-tuning before testing to only optimize the prompt parameters.
Table~\ref{tab:zeroshot} presents the comparison results of our POV and other methods.
Specifically, we compare POV with the vanilla transformer model MViT~\cite{fan2021multiscale} and several domain adaptation methods, such as AaD~\cite{yang2022attracting}, SHOT~\cite{liang2020we} and Tent~\cite{wang2021tent}.
These methods are initialized from our pre-trained model and fine-tune the backbones, while we only fine-tune prompts and keep the backbone frozen.
The results show that POV significantly outperforms other methods by a large margin, achieving at least \expbf{2.65} and \expbf{3.51} top-1 action accuracy improvement on Assembly101 and H2O, respectively.
Surprisingly, the domain adaptation methods perform poorly on the Ego-HOI-XView task, with significantly worse results compared to MViT-3rd, which does not use egocentric data.
We believe the reason is that the domain gap in Ego-HOI-XView is much larger than in traditional domain adaptation tasks due to differences in the human-body pose, object shapes, and potential occlusion caused by camera shifts. 
As a result, these methods often overfit when fine-tuning models on egocentric data.
In contrast, our POV achieves significant improvements, proving the effectiveness of prompt tuning and the essence of a frozen backbone for Ego-HOI-XView.

\noindent\textbf{Few-shot Ego-HOI-XView Recognition. } 
Under the Few-shot setup, the model is allowed to use labels from the target view for supervised egocentric fine-tuning.
The results are presented in Table~\ref{tab: few_shot}.
First, we compare POV with EgoVLP~\cite{lin2022egocentric}, which incorporates pre-training on over 3.8M video snippets from the large-scale egocentric dataset Ego4D~\cite{grauman2022ego4d}.
In contrast, POV does not use such pre-training.
Surprisingly, despite not having access to any target-view data before egocentric fine-tuning, POV still surpasses EgoVLP~\cite{lin2022egocentric} by a considerable margin, achieving \textbf{\expbf{50.91}} and \textbf{\expbf{45.53}} in top-1 verb and action accuracy improvement respectively on Assembly101.
The poor performance of EgoVLP once again demonstrates the need for careful parameter tuning in adapting to the egocentric view in the Ego-HOI-XView task.
Otherwise, even the large-scale target-view data can be detrimental.
In addition, our proposed POV achieves significant improvements across all metrics when compared to another egocentric pre-training method, Ego-Exo.

In summary, the comparison results under both Zero-shot and Few-shot setups prove that POV has an outstanding advantage in learning view-agnostic representations from third-person scenarios. Additionally, it can quickly adapt to egocentric scenarios with minimal target-view data.

\begin{table}[t]
\caption{Effectiveness of cross-view alignment (CVA) and interactive masking prompt (IMP) for POV.}
\vspace{-8pt}
\scalebox{0.9}{
\begin{tabular}{ccc@{}|cccccc}
\toprule
\multicolumn{2}{c}{Components} && \multicolumn{6}{c}{Assemble101}\\
\midrule
 \multirow{2}{*}{CVA} & \multirow{2}{*}{IMP} && \multicolumn{2}{c}{verb} & \multicolumn{2}{c}{object} & \multicolumn{2}{c}{action}\\
&&& top-1 & top-5 & top-1 & top-5 & top-1 & top-5\\
\midrule
$\times$  & $\times$ && 23.11 & 49.39 & 14.52 & 39.22 & 5.34 & 18.80\\
$\checkmark$  & $\times$ && 23.05 & 51.16 & 16.80 & 43.76 & 6.15 & 22.73\\
$\times$  &$\checkmark$ && 28.97 & 76.14 & 16.14 & 53.79 & 6.57 & 23.13\\
$\checkmark$  &$\checkmark$ && 30.33 & 77.06 & 15.41 & 51.22 & 6.78 & 23.73\\
\bottomrule
\end{tabular}
}
\label{tab:abl_3_component}
\end{table}


\subsection{Ablation Study}
In this section, we conduct a series of ablation experiments.
If not explicitly mentioned, they are assessed on H2O and use a zero-shot Ego-HOI-XView evaluation setup.

\begin{figure*}[htb]
\centering
\resizebox{\linewidth}{!}{
\includegraphics[width=7.5cm]{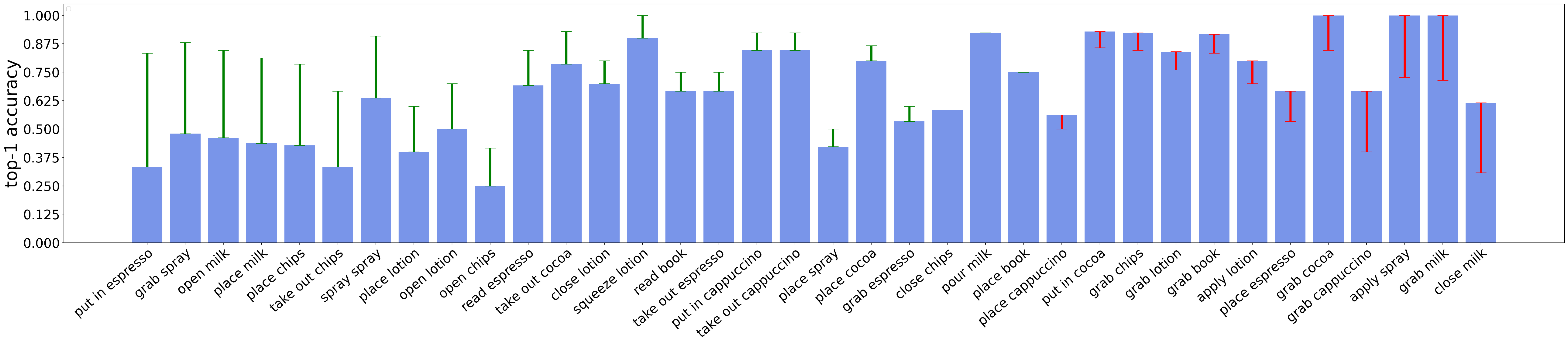}
}
\vspace{-8pt}
\caption{Illustration of per-class analysis to compare POV and MViT-3rd. The green line and red line refer to the improvement and degradation of POV over MViT-3rd respectively.}
\label{fig:class}
\end{figure*}

\noindent\textbf{(1) Effectiveness of POV to learn view-agnostic representation from third-person data. }
POV consists of two stages. 
The first stage absorbs view-agnostic knowledge from third-person data, while the second stage uses limited egocentric data.
To validate the effectiveness of the first stage, we use an evaluation setup called \textit{zero-shot HOI-XView recognition} similar to Das~\etal~\cite{das2019toyota} and is presented in Table~\ref{tab:abl_3rd}.
In this setup, a model is pre-trained on data from third-person views in $\mathcal{V}_{\rm 3rd}$, and then tested on videos of unseen third-person views that are not included in $\mathcal{V}_{\rm 3rd}$.
Compared to adapting to an egocentric view, this evaluation setup is simpler.
Our POV achieves higher verb, noun, and action accuracy than the previous state-of-the-art 3DTRL~\cite{shang2022learning}. 
This demonstrates the robustness of POV in understanding human-object interactions from various third-person views, even when encountering previously unseen views.
To further demonstrate that POV is not only robust within third-person views but also contributes to egocentric scenario learning, we introduce a new evaluation setup called \textit{3rd-to-ego HOI recognition}, as presented in Table~\ref{tab:abl_2_3rd-to-ego}.
This setup requires the model to predict human-object interactions in egocentric videos, without fine-tuning any parameters after pre-training on $D_{\rm 3rd}$.
The results show that pre-training on third-person data does help to improve performance in egocentric scenarios, as seen in the comparison between MviT-3rd and MViT-frozen.
And our POV w/o ego achieves the largest gain compared with other methods.


\begin{table}[t]
\caption{Ablation of different patch sizes and prompt type of interactive masking prompt on H2O.}
\vspace{-8pt}
\scalebox{0.9}{
\begin{tabular}{c|c|cccccc}
\toprule
\multirow{2}{*}{size} & \multirow{2}{*}{type}& \multicolumn{2}{c}{verb} & \multicolumn{2}{c}{object} & \multicolumn{2}{c}{action}\\
&& top-1 & top-5 & top-1 & top-5 & top-1 & top-5\\
\midrule
\multirow{2}{*}{1 $\times$ 1} & soft & 41.40 & 83.00 & 75.39 & 92.77 & 37.30 & 72.26\\
& hard & 35.93 & 78.32 & 72.46 & 93.94 & 33.39 & 72.85 \\
\midrule
\multirow{2}{*}{20 $\times$ 20} & soft & 44.33 & 84.96 & 80.07 & 95.11 & 41.01 & 79.29\\
& hard & 47.85 & 87.10 & 78.12 & 94.72 & 42.77 & 81.25 \\
\bottomrule
\end{tabular}}
\label{tab:interaction_prompt}
\end{table}

\begin{table}[t]
\caption{Ablation of different types of view-aware prompt: model, block and layer levels.
We report top-5 accuracy.}
\vspace{-8pt}
\scalebox{0.9}{
\begin{tabular}{c|cccc@{}|ccc}
\toprule
&\multicolumn{3}{c}{Assembly101} &&
\multicolumn{3}{c}{H2O}\\
\midrule
\multirow{1}{*}{type} & 
verb & object & action &&verb & object &action \\
\midrule
model & 73.76 & 52.16 & 20.97 && 83.00 & 93.55 & 76.17\\
block & 76.14 & 53.79 & 23.13 && 84.76 & 91.60 & 76.36 \\
layer & 73.56 & 53.66 & 18.83 && 82.42 & 90.62 & 76.17 \\
\bottomrule
\end{tabular}
}
\label{tab:view_prompt}
\end{table}

\noindent\textbf{(2) Effectiveness of egocentric prompt tuning. }
When we compare the top-1 results of POV in Table~\ref{tab: few_shot} and POV w/o ego in Table~\ref{tab:abl_2_3rd-to-ego}, we can see that supervised egocentric fine-tuning plays a significant role in improving performance.
Specifically, on Assembly101 and H2O, it leads to top-1 action accuracy improvement of \expbf{39.28} and \expbf{29.69}, respectively.
Self-supervised egocentric fine-tuning also contributes to a stable but less significant improvement, with top-5 accuracy improvement of \expbf{4.17} and \expbf{3.12} on these two datasets.
Overall, these results suggest that with only limited labeled egocentric data or even without labeled egocentric data, egocentric fine-tuning can be helpful for POV to improve performance.

\noindent\textbf{(3) Effectiveness of important components of POV third-person pre-training. } 
In this experiment, we aim to validate two components of POV: cross-view alignment (CAV) and interactive masking prompts (IMP). 
As shown in Table~\ref{tab:abl_3_component}, both components are crucial for improving action-aware representation learning.
However, we observe that when we use both components together, the accuracy for verbs and actions steadily increases, while the accuracy for objects drops.
This seeming contradiction may be due to the tension between generalization and discrimination: CVA contributes to generalizing to views that the model has not seen before, while IMP helps the model to focus on the visual cues within specific views.
Despite this trade-off, the final results are still promising, yielding top-5 accuracy improvement of \expbf{27.67} for verbs, \expbf{12} for nouns, and \expbf{4.93} for actions compared to the baseline.

\noindent\textbf{(4) Different designs of interactive masking prompt. }
In Table~\ref{tab:interaction_prompt}, we examine how different patch sizes and prompt types of interactive masking prompts affect performance. 
The results suggest that optimal performance is achieved with either a proper mask size and hard prompts, or a small patch size and soft prompts.
We conclude from Row 1 and 2 that soft prompts are effective with a small mask size because they adaptively adjust the knowledge learned from hands and objects.
Since there may be only one hand operating an object, soft prompts can provide more indication of the in-contact hand and object, while giving less attention to the other hand. 
Rows 2 and 4 reveal that an appropriate mask size for hard prompts has a positive impact on identifying the HOI region.
This is due to the relatively small size of the hand-object interaction region compared to the entire frame.
Larger patch sizes provide more overlapping and reinforce the HOI region information.

\noindent\textbf{(5) Different design of view-aware prompt .}
In Table ~\ref{tab:view_prompt}, we examine the impact of view-aware prompts by considering three variants of prompts: model level, block level, and layer level prompts. 
From the model level to the block level and then to the layer level, the prompts represent the view information at different scales of depth from single-scale to multi-scale.
The results show that the block-level prompts achieve the best performance, while the model-level and layer-level prompts perform worse.
Based on these results, we conclude that using a shallow prompt on the block level is more effective for learning view-agnostic representations at a high level. 
In contrast, using deep prompts on the layer level tends to overfit on uninformative noise rather than useful HOI information.

\begin{figure}[t]
\centering
\resizebox{\linewidth}{!}{
\includegraphics[width=8cm]{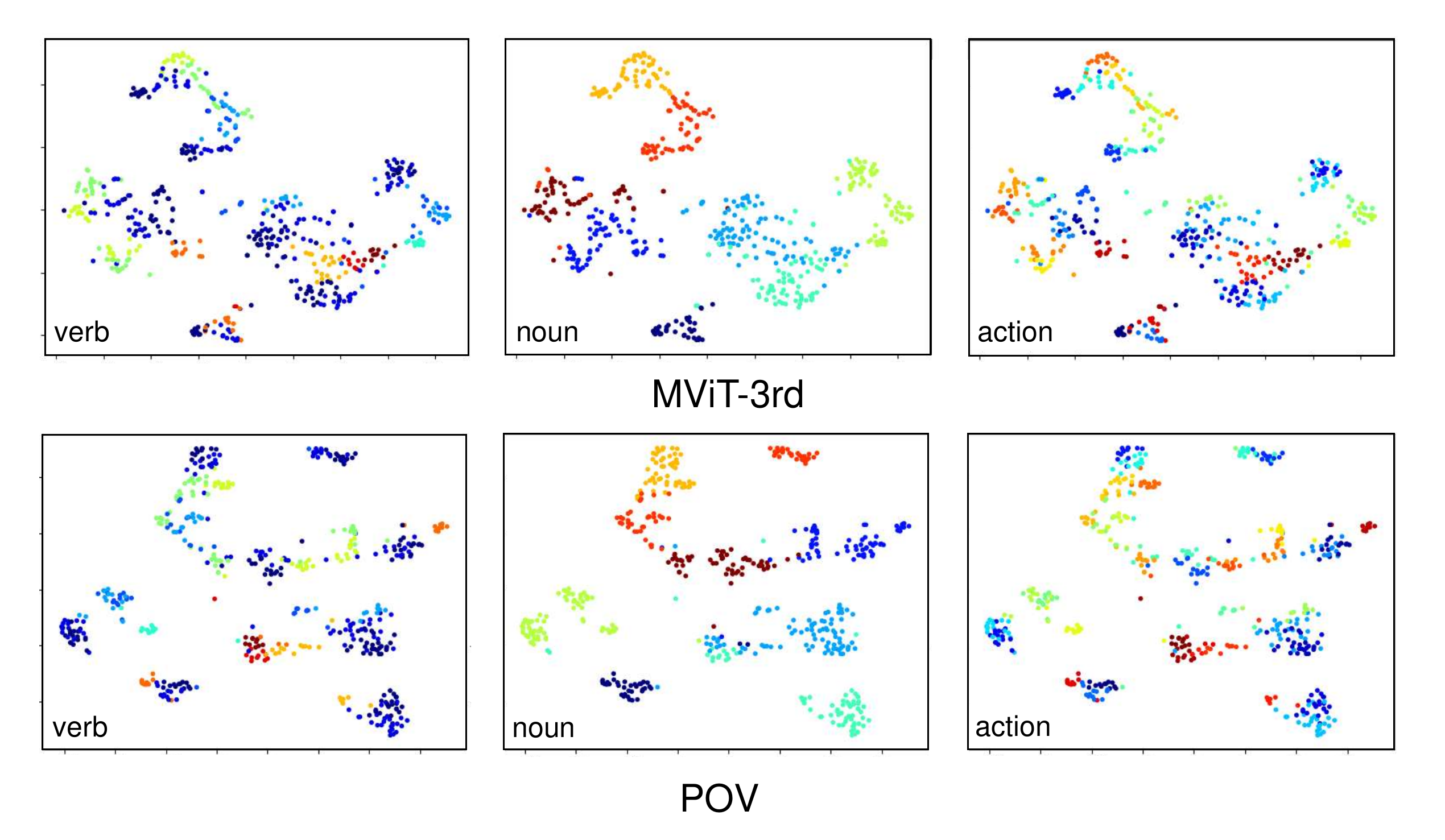}
}
\vspace{-12pt}
\caption{Illustration of t-SNE analysis of egocentric features, where each color represents one unique class.
}
\label{fig:tsne_version2}
\end{figure}

\begin{table}[t]
\caption{Computation efficiency analysis of learnable parameters (params), GPU memory consumption (GPU mem) of different tuning methods. We report top-5 accuracy.}
\vspace{-8pt}
\begin{tabular}{c|cccc}
\toprule
&AaD~\cite{yang2022attracting} & SHOT~\cite{liang2020we} & Tent~\cite{wang2021tent} & POV\\
\midrule
params/M & 34.36 & 34.21 & 0.034 & 0.006 \\
GPU mem/M & 6054 & 6052 & 4842 & 4840\\
top-5 acc/\%& 14.20 & 18.90 & 23.87 & 27.90 \\
\bottomrule
\end{tabular}
\label{tab:param}
\end{table}
\subsection{Computation Efficiency Analysis}
To evaluate the computational cost of various tuning methods including AaD (tuning the whole structure), SHOT (tuning the whole structure except the classification head), Tent (tuning only the normalization layer), and POV (tuning only the prompts), we compare the number of learnable parameters, memory consumption and corresponding results in top-5 accuracy.
The experiments are conducted with a batch size of 16 on a single GPU (A6000). 
According to the results in Table ~\ref{tab:param}, POV demonstrates a good balance between generalization ability and computational cost, resulting in approximately 20\% less memory consumption compared to full tuning.

\section{Qualitative Analysis}

Without specially mentioned, the qualitative analysis experiments are conducted on H2O under the few-shot Ego-HOI-XView setup.

\noindent\textbf{Per Class Analysis. }
Figure~\ref{fig:class} displays the per-class accuracy comparison between POV and MViT-3rd.
The figure shows that our POV brings significant improvement in most HOI categories, except for verbs like "grab" or "apply" and their combinations with nouns.
We argue that these HOI classes have ambiguous definitions and are difficult to distinguish.
Despite these challenges, the class-wise accuracy results demonstrate the robustness merit of our POV towards the Ego-HOI-XView task.

\begin{figure}[t]
\centering
\resizebox{\linewidth}{!}{
\includegraphics[width=32\linewidth]{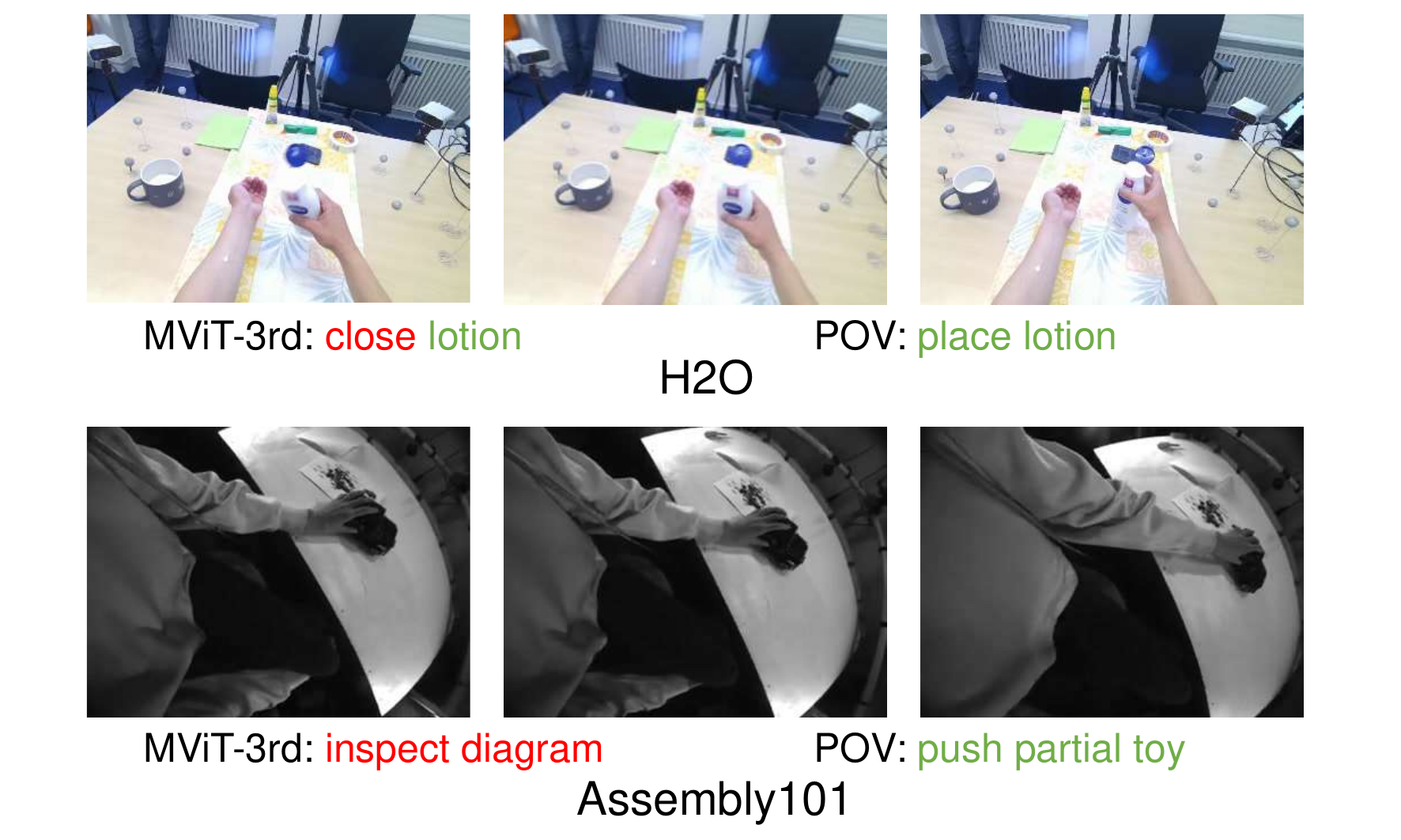}
}
\vspace{-10pt}
\caption{Illustration of egocentric prediction examples.}
\label{fig:quanlitative}
\end{figure}

\noindent\textbf{Feature Cluster Analysis. }
To analyze the feature cluster in Ego-HOI-XView, we visualize the video representations of verb, noun, and action classes in Figure~\ref{fig:tsne_version2}.
When comparing POV with the baseline MViT-3rd, we observe that video embeddings are more compact within classes and scattered between classes.
This suggests that POV learns better representation for HOI understanding, despite the view gap between pre-training third-person views and downstream egocentric views.
Furthermore, given certain classes, the spatial representation exhibits more centralized clusters compared to distributed motion representation, providing evidence to suggest that verb recognition in Ego-HOI-XView is more fine-grained than noun recognition.

\noindent\textbf{Prediction Examples. }
To better visualize the improvement brought by our POV, Figure~\ref{fig:quanlitative} provides two prediction examples from both H2O and Assembly101 datasets. 
Our POV not only correctly recognizes the fine-grained action classes ``place lotion'' in H2O, which is misclassified as ``close lotion'' by MViT-3rd, but also improves recognition results in monochrome environments in Assembly101, which shows its potential in robotic vision.

\section{CONCLUSION}
In this paper, we propose the POV framework to address the egocentric hand-object interaction understanding task.  
Inspired by the human learning process via observation, we first pre-train POV on third-person videos with two learning objectives: action understanding and view-agnostic tuning. 
We then optionally fine-tune POV on egocentric videos to improve view adaptation.
To learn fine-grained action and view-agnostic knowledge, we use a prompt-oriented vision transformer and design two types of visual prompts for view adaptation.
Extensive experiments under different evaluation setups prove the effectiveness and efficiency of POV, outperforming existing approaches.

\noindent\textbf{Discussion. }
It's worth noting that generating view labels for videos in $D_{\rm 3rd}$ is straightforward with the help of an off-the-shelf view classifier similar to Zhou~\etal~\cite{lingtao2019object} but towards the human body.
Therefore, by expanding the pre-training data with view labels, we can expect to achieve better view-agnostic representation.
We provide this suggestion for the future direction.

\begin{acks}
This work was partially supported by the National Natural Science Foundation of China (No. 62072462) and the National Key  R\&D  Program  of  China  (No.2020AAA0108600).  
\end{acks}

\bibliographystyle{ACM-Reference-Format}
\balance
\bibliography{main}


\begin{thebibliography}{69}


\ifx \showCODEN    \undefined \def \showCODEN     #1{\unskip}     \fi
\ifx \showDOI      \undefined \def \showDOI       #1{#1}\fi
\ifx \showISBNx    \undefined \def \showISBNx     #1{\unskip}     \fi
\ifx \showISBNxiii \undefined \def \showISBNxiii  #1{\unskip}     \fi
\ifx \showISSN     \undefined \def \showISSN      #1{\unskip}     \fi
\ifx \showLCCN     \undefined \def \showLCCN      #1{\unskip}     \fi
\ifx \shownote     \undefined \def \shownote      #1{#1}          \fi
\ifx \showarticletitle \undefined \def \showarticletitle #1{#1}   \fi
\ifx \showURL      \undefined \def \showURL       {\relax}        \fi
\providecommand\bibfield[2]{#2}
\providecommand\bibinfo[2]{#2}
\providecommand\natexlab[1]{#1}
\providecommand\showeprint[2][]{arXiv:#2}

\bibitem[Arnab et~al\mbox{.}(2021)]%
        {arnab2021vivit}
\bibfield{author}{\bibinfo{person}{Anurag Arnab}, \bibinfo{person}{Mostafa
  Dehghani}, \bibinfo{person}{Georg Heigold}, \bibinfo{person}{Chen Sun},
  \bibinfo{person}{Mario Lu{\v{c}}i{\'c}}, {and} \bibinfo{person}{Cordelia
  Schmid}.} \bibinfo{year}{2021}\natexlab{}.
\newblock \showarticletitle{Vivit: A video vision transformer}. In
  \bibinfo{booktitle}{\emph{Proceedings of the IEEE/CVF international
  conference on computer vision}}. \bibinfo{pages}{6836--6846}.
\newblock


\bibitem[Carreira and Zisserman(2017)]%
        {carreira2017quo}
\bibfield{author}{\bibinfo{person}{Joao Carreira} {and} \bibinfo{person}{Andrew
  Zisserman}.} \bibinfo{year}{2017}\natexlab{}.
\newblock \showarticletitle{Quo vadis, action recognition? a new model and the
  kinetics dataset}. In \bibinfo{booktitle}{\emph{proceedings of the IEEE
  Conference on Computer Vision and Pattern Recognition}}.
  \bibinfo{pages}{6299--6308}.
\newblock


\bibitem[Choi et~al\mbox{.}(2020)]%
        {Choi_2020_WACV}
\bibfield{author}{\bibinfo{person}{Jinwoo Choi}, \bibinfo{person}{Gaurav
  Sharma}, \bibinfo{person}{Manmohan Chandraker}, {and}
  \bibinfo{person}{Jia-Bin Huang}.} \bibinfo{year}{2020}\natexlab{}.
\newblock \showarticletitle{Unsupervised and Semi-Supervised Domain Adaptation
  for Action Recognition from Drones}. In \bibinfo{booktitle}{\emph{The IEEE
  Winter Conference on Applications of Computer Vision (WACV)}}.
\newblock


\bibitem[Damen et~al\mbox{.}(2018)]%
        {damen2018scaling}
\bibfield{author}{\bibinfo{person}{Dima Damen}, \bibinfo{person}{Hazel
  Doughty}, \bibinfo{person}{Giovanni~Maria Farinella}, \bibinfo{person}{Sanja
  Fidler}, \bibinfo{person}{Antonino Furnari}, \bibinfo{person}{Evangelos
  Kazakos}, \bibinfo{person}{Davide Moltisanti}, \bibinfo{person}{Jonathan
  Munro}, \bibinfo{person}{Toby Perrett}, \bibinfo{person}{Will Price},
  {et~al\mbox{.}}} \bibinfo{year}{2018}\natexlab{}.
\newblock \showarticletitle{Scaling egocentric vision: The epic-kitchens
  dataset}. In \bibinfo{booktitle}{\emph{Proceedings of the European Conference
  on Computer Vision (ECCV)}}. \bibinfo{pages}{720--736}.
\newblock


\bibitem[Damen et~al\mbox{.}(2022)]%
        {damen2022rescaling}
\bibfield{author}{\bibinfo{person}{Dima Damen}, \bibinfo{person}{Hazel
  Doughty}, \bibinfo{person}{Giovanni~Maria Farinella},
  \bibinfo{person}{Antonino Furnari}, \bibinfo{person}{Evangelos Kazakos},
  \bibinfo{person}{Jian Ma}, \bibinfo{person}{Davide Moltisanti},
  \bibinfo{person}{Jonathan Munro}, \bibinfo{person}{Toby Perrett},
  \bibinfo{person}{Will Price}, {et~al\mbox{.}}}
  \bibinfo{year}{2022}\natexlab{}.
\newblock \showarticletitle{Rescaling egocentric vision: Collection, pipeline
  and challenges for epic-kitchens-100}.
\newblock \bibinfo{journal}{\emph{International Journal of Computer Vision}}
  (\bibinfo{year}{2022}), \bibinfo{pages}{1--23}.
\newblock


\bibitem[Das et~al\mbox{.}(2019)]%
        {das2019toyota}
\bibfield{author}{\bibinfo{person}{Srijan Das}, \bibinfo{person}{Rui Dai},
  \bibinfo{person}{Michal Koperski}, \bibinfo{person}{Luca Minciullo},
  \bibinfo{person}{Lorenzo Garattoni}, \bibinfo{person}{Francois Bremond},
  {and} \bibinfo{person}{Gianpiero Francesca}.}
  \bibinfo{year}{2019}\natexlab{}.
\newblock \showarticletitle{Toyota smarthome: Real-world activities of daily
  living}. In \bibinfo{booktitle}{\emph{Proceedings of the IEEE/CVF
  international conference on computer vision}}. \bibinfo{pages}{833--842}.
\newblock


\bibitem[Das and Ryoo(2023)]%
        {das2023viewclr}
\bibfield{author}{\bibinfo{person}{Srijan Das} {and} \bibinfo{person}{Michael~S
  Ryoo}.} \bibinfo{year}{2023}\natexlab{}.
\newblock \showarticletitle{ViewCLR: Learning Self-supervised Video
  Representation for Unseen Viewpoints}. In
  \bibinfo{booktitle}{\emph{Proceedings of the IEEE/CVF Winter Conference on
  Applications of Computer Vision}}. \bibinfo{pages}{5573--5583}.
\newblock


\bibitem[Deng et~al\mbox{.}(2009)]%
        {deng2009imagenet}
\bibfield{author}{\bibinfo{person}{Jia Deng}, \bibinfo{person}{Wei Dong},
  \bibinfo{person}{Richard Socher}, \bibinfo{person}{Li-Jia Li},
  \bibinfo{person}{Kai Li}, {and} \bibinfo{person}{Li Fei-Fei}.}
  \bibinfo{year}{2009}\natexlab{}.
\newblock \showarticletitle{Imagenet: A large-scale hierarchical image
  database}. In \bibinfo{booktitle}{\emph{2009 IEEE conference on computer
  vision and pattern recognition}}. Ieee, \bibinfo{pages}{248--255}.
\newblock


\bibitem[Escorcia et~al\mbox{.}(2022)]%
        {escorcia2022sos}
\bibfield{author}{\bibinfo{person}{Victor Escorcia}, \bibinfo{person}{Ricardo
  Guerrero}, \bibinfo{person}{Xiatian Zhu}, {and} \bibinfo{person}{Brais
  Martinez}.} \bibinfo{year}{2022}\natexlab{}.
\newblock \showarticletitle{SOS! Self-supervised Learning over Sets of Handled
  Objects in Egocentric Action Recognition}. In
  \bibinfo{booktitle}{\emph{Computer Vision--ECCV 2022: 17th European
  Conference, Tel Aviv, Israel, October 23--27, 2022, Proceedings, Part XIII}}.
  Springer, \bibinfo{pages}{604--620}.
\newblock


\bibitem[Fan et~al\mbox{.}(2021)]%
        {fan2021multiscale}
\bibfield{author}{\bibinfo{person}{Haoqi Fan}, \bibinfo{person}{Bo Xiong},
  \bibinfo{person}{Karttikeya Mangalam}, \bibinfo{person}{Yanghao Li},
  \bibinfo{person}{Zhicheng Yan}, \bibinfo{person}{Jitendra Malik}, {and}
  \bibinfo{person}{Christoph Feichtenhofer}.} \bibinfo{year}{2021}\natexlab{}.
\newblock \showarticletitle{Multiscale vision transformers}. In
  \bibinfo{booktitle}{\emph{Proceedings of the IEEE/CVF International
  Conference on Computer Vision}}. \bibinfo{pages}{6824--6835}.
\newblock


\bibitem[Farhadi and Tabrizi(2008)]%
        {farhadi2008learning}
\bibfield{author}{\bibinfo{person}{Ali Farhadi} {and}
  \bibinfo{person}{Mostafa~Kamali Tabrizi}.} \bibinfo{year}{2008}\natexlab{}.
\newblock \showarticletitle{Learning to recognize activities from the wrong
  view point}. In \bibinfo{booktitle}{\emph{Computer Vision--ECCV 2008: 10th
  European Conference on Computer Vision, Marseille, France, October 12-18,
  2008, Proceedings, Part I 10}}. Springer, \bibinfo{pages}{154--166}.
\newblock


\bibitem[Frankle and Carbin(2018)]%
        {frankle2018lottery}
\bibfield{author}{\bibinfo{person}{Jonathan Frankle} {and}
  \bibinfo{person}{Michael Carbin}.} \bibinfo{year}{2018}\natexlab{}.
\newblock \showarticletitle{The lottery ticket hypothesis: Finding sparse,
  trainable neural networks}.
\newblock \bibinfo{journal}{\emph{arXiv preprint arXiv:1803.03635}}
  (\bibinfo{year}{2018}).
\newblock


\bibitem[Furnari and Farinella(2020)]%
        {furnari2020rolling}
\bibfield{author}{\bibinfo{person}{Antonino Furnari} {and}
  \bibinfo{person}{Giovanni~Maria Farinella}.} \bibinfo{year}{2020}\natexlab{}.
\newblock \showarticletitle{Rolling-unrolling lstms for action anticipation
  from first-person video}.
\newblock \bibinfo{journal}{\emph{IEEE transactions on pattern analysis and
  machine intelligence}} \bibinfo{volume}{43}, \bibinfo{number}{11}
  (\bibinfo{year}{2020}), \bibinfo{pages}{4021--4036}.
\newblock


\bibitem[Gao et~al\mbox{.}(2022)]%
        {gao2022visual}
\bibfield{author}{\bibinfo{person}{Yunhe Gao}, \bibinfo{person}{Xingjian Shi},
  \bibinfo{person}{Yi Zhu}, \bibinfo{person}{Hao Wang},
  \bibinfo{person}{Zhiqiang Tang}, \bibinfo{person}{Xiong Zhou},
  \bibinfo{person}{Mu Li}, {and} \bibinfo{person}{Dimitris~N Metaxas}.}
  \bibinfo{year}{2022}\natexlab{}.
\newblock \showarticletitle{Visual Prompt Tuning for Test-time Domain
  Adaptation}.
\newblock \bibinfo{journal}{\emph{arXiv preprint arXiv:2210.04831}}
  (\bibinfo{year}{2022}).
\newblock


\bibitem[Goyal et~al\mbox{.}(2017)]%
        {goyal2017something}
\bibfield{author}{\bibinfo{person}{Raghav Goyal}, \bibinfo{person}{Samira
  Ebrahimi~Kahou}, \bibinfo{person}{Vincent Michalski}, \bibinfo{person}{Joanna
  Materzynska}, \bibinfo{person}{Susanne Westphal}, \bibinfo{person}{Heuna
  Kim}, \bibinfo{person}{Valentin Haenel}, \bibinfo{person}{Ingo Fruend},
  \bibinfo{person}{Peter Yianilos}, \bibinfo{person}{Moritz Mueller-Freitag},
  {et~al\mbox{.}}} \bibinfo{year}{2017}\natexlab{}.
\newblock \showarticletitle{The" something something" video database for
  learning and evaluating visual common sense}. In
  \bibinfo{booktitle}{\emph{Proceedings of the IEEE international conference on
  computer vision}}. \bibinfo{pages}{5842--5850}.
\newblock


\bibitem[Grauman et~al\mbox{.}(2022)]%
        {grauman2022ego4d}
\bibfield{author}{\bibinfo{person}{Kristen Grauman}, \bibinfo{person}{Andrew
  Westbury}, \bibinfo{person}{Eugene Byrne}, \bibinfo{person}{Zachary Chavis},
  \bibinfo{person}{Antonino Furnari}, \bibinfo{person}{Rohit Girdhar},
  \bibinfo{person}{Jackson Hamburger}, \bibinfo{person}{Hao Jiang},
  \bibinfo{person}{Miao Liu}, \bibinfo{person}{Xingyu Liu}, {et~al\mbox{.}}}
  \bibinfo{year}{2022}\natexlab{}.
\newblock \showarticletitle{Ego4d: Around the world in 3,000 hours of
  egocentric video}. In \bibinfo{booktitle}{\emph{Proceedings of the IEEE/CVF
  Conference on Computer Vision and Pattern Recognition}}.
  \bibinfo{pages}{18995--19012}.
\newblock


\bibitem[Han et~al\mbox{.}(2020)]%
        {han2020megatrack}
\bibfield{author}{\bibinfo{person}{Shangchen Han}, \bibinfo{person}{Beibei
  Liu}, \bibinfo{person}{Randi Cabezas}, \bibinfo{person}{Christopher~D Twigg},
  \bibinfo{person}{Peizhao Zhang}, \bibinfo{person}{Jeff Petkau},
  \bibinfo{person}{Tsz-Ho Yu}, \bibinfo{person}{Chun-Jung Tai},
  \bibinfo{person}{Muzaffer Akbay}, \bibinfo{person}{Zheng Wang},
  {et~al\mbox{.}}} \bibinfo{year}{2020}\natexlab{}.
\newblock \showarticletitle{MEgATrack: monochrome egocentric articulated
  hand-tracking for virtual reality}.
\newblock \bibinfo{journal}{\emph{ACM Transactions on Graphics (ToG)}}
  \bibinfo{volume}{39}, \bibinfo{number}{4} (\bibinfo{year}{2020}),
  \bibinfo{pages}{87--1}.
\newblock


\bibitem[Herath et~al\mbox{.}(2017)]%
        {herath2017going}
\bibfield{author}{\bibinfo{person}{Samitha Herath}, \bibinfo{person}{Mehrtash
  Harandi}, {and} \bibinfo{person}{Fatih Porikli}.}
  \bibinfo{year}{2017}\natexlab{}.
\newblock \showarticletitle{Going deeper into action recognition: A survey}.
\newblock \bibinfo{journal}{\emph{Image and vision computing}}
  \bibinfo{volume}{60} (\bibinfo{year}{2017}), \bibinfo{pages}{4--21}.
\newblock


\bibitem[Jaderberg et~al\mbox{.}(2015)]%
        {jaderberg2015spatial}
\bibfield{author}{\bibinfo{person}{Max Jaderberg}, \bibinfo{person}{Karen
  Simonyan}, \bibinfo{person}{Andrew Zisserman}, {et~al\mbox{.}}}
  \bibinfo{year}{2015}\natexlab{}.
\newblock \showarticletitle{Spatial transformer networks}.
\newblock \bibinfo{journal}{\emph{Advances in neural information processing
  systems}}  \bibinfo{volume}{28} (\bibinfo{year}{2015}).
\newblock


\bibitem[Jangir et~al\mbox{.}(2022)]%
        {jangir2022look}
\bibfield{author}{\bibinfo{person}{Rishabh Jangir}, \bibinfo{person}{Nicklas
  Hansen}, \bibinfo{person}{Sambaran Ghosal}, \bibinfo{person}{Mohit Jain},
  {and} \bibinfo{person}{Xiaolong Wang}.} \bibinfo{year}{2022}\natexlab{}.
\newblock \showarticletitle{Look closer: Bridging egocentric and third-person
  views with transformers for robotic manipulation}.
\newblock \bibinfo{journal}{\emph{IEEE Robotics and Automation Letters}}
  \bibinfo{volume}{7}, \bibinfo{number}{2} (\bibinfo{year}{2022}),
  \bibinfo{pages}{3046--3053}.
\newblock


\bibitem[Ji et~al\mbox{.}(2019)]%
        {ji2019attention}
\bibfield{author}{\bibinfo{person}{Yanli Ji}, \bibinfo{person}{Feixiang Xu},
  \bibinfo{person}{Yang Yang}, \bibinfo{person}{Ning Xie},
  \bibinfo{person}{Heng~Tao Shen}, {and} \bibinfo{person}{Tatsuya Harada}.}
  \bibinfo{year}{2019}\natexlab{}.
\newblock \showarticletitle{Attention transfer (ANT) network for view-invariant
  action recognition}. In \bibinfo{booktitle}{\emph{Proceedings of the 27th ACM
  International Conference on Multimedia}}. \bibinfo{pages}{574--582}.
\newblock


\bibitem[Jia et~al\mbox{.}(2022a)]%
        {DBLP:conf/eccv/VPT22}
\bibfield{author}{\bibinfo{person}{Menglin Jia}, \bibinfo{person}{Luming Tang},
  \bibinfo{person}{Bor{-}Chun Chen}, \bibinfo{person}{Claire Cardie},
  \bibinfo{person}{Serge~J. Belongie}, \bibinfo{person}{Bharath Hariharan},
  {and} \bibinfo{person}{Ser{-}Nam Lim}.} \bibinfo{year}{2022}\natexlab{a}.
\newblock \showarticletitle{Visual Prompt Tuning}. In
  \bibinfo{booktitle}{\emph{Computer Vision - {ECCV} 2022 - 17th European
  Conference, Tel Aviv, Israel, October 23-27, 2022, Proceedings, Part
  {XXXIII}}} \emph{(\bibinfo{series}{Lecture Notes in Computer Science},
  Vol.~\bibinfo{volume}{13693})}, \bibfield{editor}{\bibinfo{person}{Shai
  Avidan}, \bibinfo{person}{Gabriel~J. Brostow}, \bibinfo{person}{Moustapha
  Ciss{\'{e}}}, \bibinfo{person}{Giovanni~Maria Farinella}, {and}
  \bibinfo{person}{Tal Hassner}} (Eds.). \bibinfo{publisher}{Springer},
  \bibinfo{pages}{709--727}.
\newblock
\urldef\tempurl%
\url{https://doi.org/10.1007/978-3-031-19827-4\_41}
\showDOI{\tempurl}


\bibitem[Jia et~al\mbox{.}(2022b)]%
        {jia2022visual}
\bibfield{author}{\bibinfo{person}{Menglin Jia}, \bibinfo{person}{Luming Tang},
  \bibinfo{person}{Bor-Chun Chen}, \bibinfo{person}{Claire Cardie},
  \bibinfo{person}{Serge Belongie}, \bibinfo{person}{Bharath Hariharan}, {and}
  \bibinfo{person}{Ser-Nam Lim}.} \bibinfo{year}{2022}\natexlab{b}.
\newblock \showarticletitle{Visual prompt tuning}. In
  \bibinfo{booktitle}{\emph{Computer Vision--ECCV 2022: 17th European
  Conference, Tel Aviv, Israel, October 23--27, 2022, Proceedings, Part
  XXXIII}}. Springer, \bibinfo{pages}{709--727}.
\newblock


\bibitem[Jiang et~al\mbox{.}(2020)]%
        {jiang2020can}
\bibfield{author}{\bibinfo{person}{Zhengbao Jiang}, \bibinfo{person}{Frank~F
  Xu}, \bibinfo{person}{Jun Araki}, {and} \bibinfo{person}{Graham Neubig}.}
  \bibinfo{year}{2020}\natexlab{}.
\newblock \showarticletitle{How can we know what language models know?}
\newblock \bibinfo{journal}{\emph{Transactions of the Association for
  Computational Linguistics}}  \bibinfo{volume}{8} (\bibinfo{year}{2020}),
  \bibinfo{pages}{423--438}.
\newblock


\bibitem[Kay et~al\mbox{.}(2017)]%
        {kay2017kinetics}
\bibfield{author}{\bibinfo{person}{Will Kay}, \bibinfo{person}{Joao Carreira},
  \bibinfo{person}{Karen Simonyan}, \bibinfo{person}{Brian Zhang},
  \bibinfo{person}{Chloe Hillier}, \bibinfo{person}{Sudheendra
  Vijayanarasimhan}, \bibinfo{person}{Fabio Viola}, \bibinfo{person}{Tim
  Green}, \bibinfo{person}{Trevor Back}, \bibinfo{person}{Paul Natsev},
  {et~al\mbox{.}}} \bibinfo{year}{2017}\natexlab{}.
\newblock \showarticletitle{The kinetics human action video dataset}.
\newblock \bibinfo{journal}{\emph{arXiv preprint arXiv:1705.06950}}
  (\bibinfo{year}{2017}).
\newblock


\bibitem[Kazakos et~al\mbox{.}(2019)]%
        {kazakos2019epic}
\bibfield{author}{\bibinfo{person}{Evangelos Kazakos}, \bibinfo{person}{Arsha
  Nagrani}, \bibinfo{person}{Andrew Zisserman}, {and} \bibinfo{person}{Dima
  Damen}.} \bibinfo{year}{2019}\natexlab{}.
\newblock \showarticletitle{Epic-fusion: Audio-visual temporal binding for
  egocentric action recognition}. In \bibinfo{booktitle}{\emph{Proceedings of
  the IEEE/CVF International Conference on Computer Vision}}.
  \bibinfo{pages}{5492--5501}.
\newblock


\bibitem[Kim et~al\mbox{.}(2019)]%
        {kim2019eyes}
\bibfield{author}{\bibinfo{person}{Daekyum Kim},
  \bibinfo{person}{Brian~Byunghyun Kang}, \bibinfo{person}{Kyu~Bum Kim},
  \bibinfo{person}{Hyungmin Choi}, \bibinfo{person}{Jeesoo Ha},
  \bibinfo{person}{Kyu-Jin Cho}, {and} \bibinfo{person}{Sungho Jo}.}
  \bibinfo{year}{2019}\natexlab{}.
\newblock \showarticletitle{Eyes are faster than hands: A soft wearable robot
  learns user intention from the egocentric view}.
\newblock \bibinfo{journal}{\emph{Science Robotics}} \bibinfo{volume}{4},
  \bibinfo{number}{26} (\bibinfo{year}{2019}), \bibinfo{pages}{eaav2949}.
\newblock


\bibitem[Kong and Fu(2022)]%
        {kong2022human}
\bibfield{author}{\bibinfo{person}{Yu Kong} {and} \bibinfo{person}{Yun Fu}.}
  \bibinfo{year}{2022}\natexlab{}.
\newblock \showarticletitle{Human action recognition and prediction: A survey}.
\newblock \bibinfo{journal}{\emph{International Journal of Computer Vision}}
  \bibinfo{volume}{130}, \bibinfo{number}{5} (\bibinfo{year}{2022}),
  \bibinfo{pages}{1366--1401}.
\newblock


\bibitem[Kwon et~al\mbox{.}(2021)]%
        {kwon2021h2o}
\bibfield{author}{\bibinfo{person}{Taein Kwon}, \bibinfo{person}{Bugra Tekin},
  \bibinfo{person}{Jan St{\"u}hmer}, \bibinfo{person}{Federica Bogo}, {and}
  \bibinfo{person}{Marc Pollefeys}.} \bibinfo{year}{2021}\natexlab{}.
\newblock \showarticletitle{H2o: Two hands manipulating objects for first
  person interaction recognition}. In \bibinfo{booktitle}{\emph{Proceedings of
  the IEEE/CVF International Conference on Computer Vision}}.
  \bibinfo{pages}{10138--10148}.
\newblock


\bibitem[Lester et~al\mbox{.}(2021)]%
        {lester2021power}
\bibfield{author}{\bibinfo{person}{Brian Lester}, \bibinfo{person}{Rami
  Al-Rfou}, {and} \bibinfo{person}{Noah Constant}.}
  \bibinfo{year}{2021}\natexlab{}.
\newblock \showarticletitle{The power of scale for parameter-efficient prompt
  tuning}.
\newblock \bibinfo{journal}{\emph{arXiv preprint arXiv:2104.08691}}
  (\bibinfo{year}{2021}).
\newblock


\bibitem[Li and Liang(2021)]%
        {li2021prefix}
\bibfield{author}{\bibinfo{person}{Xiang~Lisa Li} {and} \bibinfo{person}{Percy
  Liang}.} \bibinfo{year}{2021}\natexlab{}.
\newblock \showarticletitle{Prefix-tuning: Optimizing continuous prompts for
  generation}.
\newblock \bibinfo{journal}{\emph{arXiv preprint arXiv:2101.00190}}
  (\bibinfo{year}{2021}).
\newblock


\bibitem[Li et~al\mbox{.}(2021a)]%
        {li2021eye}
\bibfield{author}{\bibinfo{person}{Yin Li}, \bibinfo{person}{Miao Liu}, {and}
  \bibinfo{person}{Jame Rehg}.} \bibinfo{year}{2021}\natexlab{a}.
\newblock \showarticletitle{In the eye of the beholder: Gaze and actions in
  first person video}.
\newblock \bibinfo{journal}{\emph{IEEE Transactions on Pattern Analysis and
  Machine Intelligence}} (\bibinfo{year}{2021}).
\newblock


\bibitem[Li et~al\mbox{.}(2021b)]%
        {li2021ego}
\bibfield{author}{\bibinfo{person}{Yanghao Li}, \bibinfo{person}{Tushar
  Nagarajan}, \bibinfo{person}{Bo Xiong}, {and} \bibinfo{person}{Kristen
  Grauman}.} \bibinfo{year}{2021}\natexlab{b}.
\newblock \showarticletitle{Ego-exo: Transferring visual representations from
  third-person to first-person videos}. In
  \bibinfo{booktitle}{\emph{Proceedings of the IEEE/CVF Conference on Computer
  Vision and Pattern Recognition}}. \bibinfo{pages}{6943--6953}.
\newblock


\bibitem[Li et~al\mbox{.}(2022)]%
        {li2022bevformer}
\bibfield{author}{\bibinfo{person}{Zhiqi Li}, \bibinfo{person}{Wenhai Wang},
  \bibinfo{person}{Hongyang Li}, \bibinfo{person}{Enze Xie},
  \bibinfo{person}{Chonghao Sima}, \bibinfo{person}{Tong Lu},
  \bibinfo{person}{Yu Qiao}, {and} \bibinfo{person}{Jifeng Dai}.}
  \bibinfo{year}{2022}\natexlab{}.
\newblock \showarticletitle{Bevformer: Learning bird’s-eye-view
  representation from multi-camera images via spatiotemporal transformers}. In
  \bibinfo{booktitle}{\emph{Computer Vision--ECCV 2022: 17th European
  Conference, Tel Aviv, Israel, October 23--27, 2022, Proceedings, Part IX}}.
  Springer, \bibinfo{pages}{1--18}.
\newblock


\bibitem[Liang et~al\mbox{.}(2020)]%
        {liang2020we}
\bibfield{author}{\bibinfo{person}{Jian Liang}, \bibinfo{person}{Dapeng Hu},
  {and} \bibinfo{person}{Jiashi Feng}.} \bibinfo{year}{2020}\natexlab{}.
\newblock \showarticletitle{Do we really need to access the source data? source
  hypothesis transfer for unsupervised domain adaptation}. In
  \bibinfo{booktitle}{\emph{International Conference on Machine Learning}}.
  PMLR, \bibinfo{pages}{6028--6039}.
\newblock


\bibitem[Lin et~al\mbox{.}(2022)]%
        {lin2022egocentric}
\bibfield{author}{\bibinfo{person}{Kevin~Qinghong Lin},
  \bibinfo{person}{Jinpeng Wang}, \bibinfo{person}{Mattia Soldan},
  \bibinfo{person}{Michael Wray}, \bibinfo{person}{Rui Yan},
  \bibinfo{person}{Eric~Z XU}, \bibinfo{person}{Difei Gao},
  \bibinfo{person}{Rong-Cheng Tu}, \bibinfo{person}{Wenzhe Zhao},
  \bibinfo{person}{Weijie Kong}, {et~al\mbox{.}}}
  \bibinfo{year}{2022}\natexlab{}.
\newblock \showarticletitle{Egocentric video-language pretraining}.
\newblock \bibinfo{journal}{\emph{Advances in Neural Information Processing
  Systems}}  \bibinfo{volume}{35} (\bibinfo{year}{2022}),
  \bibinfo{pages}{7575--7586}.
\newblock


\bibitem[Lingtao et~al\mbox{.}(2019)]%
        {lingtao2019object}
\bibfield{author}{\bibinfo{person}{Zhou Lingtao}, \bibinfo{person}{Fang
  Jiaojiao}, {and} \bibinfo{person}{Liu Guizhong}.}
  \bibinfo{year}{2019}\natexlab{}.
\newblock \showarticletitle{Object viewpoint classification based 3d bounding
  box estimation for autonomous vehicles}.
\newblock \bibinfo{journal}{\emph{arXiv preprint arXiv:1909.01025}}
  (\bibinfo{year}{2019}).
\newblock


\bibitem[Liu et~al\mbox{.}(2020)]%
        {liu2020ntu}
\bibfield{author}{\bibinfo{person}{Jun Liu}, \bibinfo{person}{Amir Shahroudy},
  \bibinfo{person}{Mauricio Perez}, \bibinfo{person}{Gang Wang},
  \bibinfo{person}{Ling-Yu Duan}, {and} \bibinfo{person}{Alex~C Kot}.}
  \bibinfo{year}{2020}\natexlab{}.
\newblock \showarticletitle{NTU RGB+D 120: A large-scale benchmark for 3D human
  activity understanding}.
\newblock \bibinfo{journal}{\emph{IEEE Transactions on Pattern Analysis and
  Machine Intelligence}} \bibinfo{volume}{42}, \bibinfo{number}{10}
  (\bibinfo{year}{2020}), \bibinfo{pages}{2684--2701}.
\newblock


\bibitem[Liu et~al\mbox{.}(2023)]%
        {liu2023pre}
\bibfield{author}{\bibinfo{person}{Pengfei Liu}, \bibinfo{person}{Weizhe Yuan},
  \bibinfo{person}{Jinlan Fu}, \bibinfo{person}{Zhengbao Jiang},
  \bibinfo{person}{Hiroaki Hayashi}, {and} \bibinfo{person}{Graham Neubig}.}
  \bibinfo{year}{2023}\natexlab{}.
\newblock \showarticletitle{Pre-train, prompt, and predict: A systematic survey
  of prompting methods in natural language processing}.
\newblock \bibinfo{journal}{\emph{Comput. Surveys}} \bibinfo{volume}{55},
  \bibinfo{number}{9} (\bibinfo{year}{2023}), \bibinfo{pages}{1--35}.
\newblock


\bibitem[Liu et~al\mbox{.}(2022a)]%
        {liu2022hoi4d}
\bibfield{author}{\bibinfo{person}{Yunze Liu}, \bibinfo{person}{Yun Liu},
  \bibinfo{person}{Che Jiang}, \bibinfo{person}{Kangbo Lyu},
  \bibinfo{person}{Weikang Wan}, \bibinfo{person}{Hao Shen},
  \bibinfo{person}{Boqiang Liang}, \bibinfo{person}{Zhoujie Fu},
  \bibinfo{person}{He Wang}, {and} \bibinfo{person}{Li Yi}.}
  \bibinfo{year}{2022}\natexlab{a}.
\newblock \showarticletitle{HOI4D: A 4D Egocentric Dataset for Category-Level
  Human-Object Interaction}. In \bibinfo{booktitle}{\emph{Proceedings of the
  IEEE/CVF Conference on Computer Vision and Pattern Recognition}}.
  \bibinfo{pages}{21013--21022}.
\newblock


\bibitem[Liu et~al\mbox{.}(2022b)]%
        {liu2022petr}
\bibfield{author}{\bibinfo{person}{Yingfei Liu}, \bibinfo{person}{Tiancai
  Wang}, \bibinfo{person}{Xiangyu Zhang}, {and} \bibinfo{person}{Jian Sun}.}
  \bibinfo{year}{2022}\natexlab{b}.
\newblock \showarticletitle{Petr: Position embedding transformation for
  multi-view 3d object detection}. In \bibinfo{booktitle}{\emph{Computer
  Vision--ECCV 2022: 17th European Conference, Tel Aviv, Israel, October
  23--27, 2022, Proceedings, Part XXVII}}. Springer, \bibinfo{pages}{531--548}.
\newblock


\bibitem[Lu et~al\mbox{.}(2019)]%
        {lu2019learning}
\bibfield{author}{\bibinfo{person}{Minlong Lu}, \bibinfo{person}{Danping Liao},
  {and} \bibinfo{person}{Ze-Nian Li}.} \bibinfo{year}{2019}\natexlab{}.
\newblock \showarticletitle{Learning spatiotemporal attention for egocentric
  action recognition}. In \bibinfo{booktitle}{\emph{Proceedings of the IEEE/CVF
  International Conference on Computer Vision Workshops}}.
  \bibinfo{pages}{0--0}.
\newblock


\bibitem[Miech et~al\mbox{.}(2019)]%
        {miech2019howto100m}
\bibfield{author}{\bibinfo{person}{Antoine Miech}, \bibinfo{person}{Dimitri
  Zhukov}, \bibinfo{person}{Jean-Baptiste Alayrac}, \bibinfo{person}{Makarand
  Tapaswi}, \bibinfo{person}{Ivan Laptev}, {and} \bibinfo{person}{Josef
  Sivic}.} \bibinfo{year}{2019}\natexlab{}.
\newblock \showarticletitle{Howto100m: Learning a text-video embedding by
  watching hundred million narrated video clips}. In
  \bibinfo{booktitle}{\emph{Proceedings of the IEEE/CVF International
  Conference on Computer Vision}}. \bibinfo{pages}{2630--2640}.
\newblock


\bibitem[Piergiovanni and Ryoo(2021)]%
        {piergiovanni2021recognizing}
\bibfield{author}{\bibinfo{person}{AJ Piergiovanni} {and}
  \bibinfo{person}{Michael~S Ryoo}.} \bibinfo{year}{2021}\natexlab{}.
\newblock \showarticletitle{Recognizing actions in videos from unseen
  viewpoints}. In \bibinfo{booktitle}{\emph{Proceedings of the IEEE/CVF
  Conference on Computer Vision and Pattern Recognition}}.
  \bibinfo{pages}{4124--4132}.
\newblock


\bibitem[Poppe(2010)]%
        {poppe2010survey}
\bibfield{author}{\bibinfo{person}{Ronald Poppe}.}
  \bibinfo{year}{2010}\natexlab{}.
\newblock \showarticletitle{A survey on vision-based human action recognition}.
\newblock \bibinfo{journal}{\emph{Image and vision computing}}
  \bibinfo{volume}{28}, \bibinfo{number}{6} (\bibinfo{year}{2010}),
  \bibinfo{pages}{976--990}.
\newblock


\bibitem[Qiu et~al\mbox{.}(2021)]%
        {qiu2021ego}
\bibfield{author}{\bibinfo{person}{Haonan Qiu}, \bibinfo{person}{Pan He},
  \bibinfo{person}{Shuchun Liu}, \bibinfo{person}{Weiyuan Shao},
  \bibinfo{person}{Feiyun Zhang}, \bibinfo{person}{Jiajun Wang},
  \bibinfo{person}{Liang He}, {and} \bibinfo{person}{Feng Wang}.}
  \bibinfo{year}{2021}\natexlab{}.
\newblock \showarticletitle{Ego-Deliver: A Large-Scale Dataset For Egocentric
  Video Analysis}. In \bibinfo{booktitle}{\emph{Proceedings of the 29th ACM
  International Conference on Multimedia}}. \bibinfo{pages}{1847--1855}.
\newblock


\bibitem[Rao and Shah(2001)]%
        {rao2001view}
\bibfield{author}{\bibinfo{person}{Cen Rao} {and} \bibinfo{person}{Mubarak
  Shah}.} \bibinfo{year}{2001}\natexlab{}.
\newblock \showarticletitle{View-invariance in action recognition}. In
  \bibinfo{booktitle}{\emph{Proceedings of the 2001 IEEE Computer Society
  Conference on Computer Vision and Pattern Recognition. CVPR 2001}},
  Vol.~\bibinfo{volume}{2}. IEEE, \bibinfo{pages}{II--II}.
\newblock


\bibitem[Sabater et~al\mbox{.}(2021)]%
        {sabater2021domain}
\bibfield{author}{\bibinfo{person}{Alberto Sabater}, \bibinfo{person}{I{\~n}igo
  Alonso}, \bibinfo{person}{Luis Montesano}, {and} \bibinfo{person}{Ana~C
  Murillo}.} \bibinfo{year}{2021}\natexlab{}.
\newblock \showarticletitle{Domain and view-point Agnostic hand action
  recognition}.
\newblock \bibinfo{journal}{\emph{IEEE Robotics and Automation Letters}}
  \bibinfo{volume}{6}, \bibinfo{number}{4} (\bibinfo{year}{2021}),
  \bibinfo{pages}{7823--7830}.
\newblock


\bibitem[Sener et~al\mbox{.}(2022)]%
        {sener2022assembly101}
\bibfield{author}{\bibinfo{person}{Fadime Sener}, \bibinfo{person}{Dibyadip
  Chatterjee}, \bibinfo{person}{Daniel Shelepov}, \bibinfo{person}{Kun He},
  \bibinfo{person}{Dipika Singhania}, \bibinfo{person}{Robert Wang}, {and}
  \bibinfo{person}{Angela Yao}.} \bibinfo{year}{2022}\natexlab{}.
\newblock \showarticletitle{Assembly101: A Large-Scale Multi-View Video Dataset
  for Understanding Procedural Activities}. In
  \bibinfo{booktitle}{\emph{Proceedings of the IEEE/CVF Conference on Computer
  Vision and Pattern Recognition}}. \bibinfo{pages}{21096--21106}.
\newblock


\bibitem[Shah et~al\mbox{.}(2023)]%
        {shah2023multi}
\bibfield{author}{\bibinfo{person}{Ketul Shah}, \bibinfo{person}{Anshul Shah},
  \bibinfo{person}{Chun~Pong Lau}, \bibinfo{person}{Celso~M de Melo}, {and}
  \bibinfo{person}{Rama Chellappa}.} \bibinfo{year}{2023}\natexlab{}.
\newblock \showarticletitle{Multi-View Action Recognition Using Contrastive
  Learning}. In \bibinfo{booktitle}{\emph{Proceedings of the IEEE/CVF Winter
  Conference on Applications of Computer Vision}}. \bibinfo{pages}{3381--3391}.
\newblock


\bibitem[Shan et~al\mbox{.}(2020)]%
        {shan2020understanding}
\bibfield{author}{\bibinfo{person}{Dandan Shan}, \bibinfo{person}{Jiaqi Geng},
  \bibinfo{person}{Michelle Shu}, {and} \bibinfo{person}{David~F Fouhey}.}
  \bibinfo{year}{2020}\natexlab{}.
\newblock \showarticletitle{Understanding human hands in contact at internet
  scale}. In \bibinfo{booktitle}{\emph{Proceedings of the IEEE/CVF conference
  on computer vision and pattern recognition}}. \bibinfo{pages}{9869--9878}.
\newblock


\bibitem[Shang et~al\mbox{.}(2022)]%
        {shang2022learning}
\bibfield{author}{\bibinfo{person}{Jinghuan Shang}, \bibinfo{person}{Srijan
  Das}, {and} \bibinfo{person}{Michael Ryoo}.} \bibinfo{year}{2022}\natexlab{}.
\newblock \showarticletitle{Learning viewpoint-agnostic visual representations
  by recovering tokens in 3D space}.
\newblock \bibinfo{journal}{\emph{Advances in Neural Information Processing
  Systems}}  \bibinfo{volume}{35} (\bibinfo{year}{2022}),
  \bibinfo{pages}{31031--31044}.
\newblock


\bibitem[Sharma et~al\mbox{.}(2019)]%
        {sharma2019third}
\bibfield{author}{\bibinfo{person}{Pratyusha Sharma}, \bibinfo{person}{Deepak
  Pathak}, {and} \bibinfo{person}{Abhinav Gupta}.}
  \bibinfo{year}{2019}\natexlab{}.
\newblock \showarticletitle{Third-person visual imitation learning via
  decoupled hierarchical controller}.
\newblock \bibinfo{journal}{\emph{Advances in Neural Information Processing
  Systems}}  \bibinfo{volume}{32} (\bibinfo{year}{2019}).
\newblock


\bibitem[Shen and Foroosh(2009)]%
        {shen2009view}
\bibfield{author}{\bibinfo{person}{Yuping Shen} {and} \bibinfo{person}{Hassan
  Foroosh}.} \bibinfo{year}{2009}\natexlab{}.
\newblock \showarticletitle{View-invariant action recognition from point
  triplets}.
\newblock \bibinfo{journal}{\emph{IEEE transactions on pattern analysis and
  machine intelligence}} \bibinfo{volume}{31}, \bibinfo{number}{10}
  (\bibinfo{year}{2009}), \bibinfo{pages}{1898--1905}.
\newblock


\bibitem[Shin et~al\mbox{.}(2020)]%
        {shin2020autoprompt}
\bibfield{author}{\bibinfo{person}{Taylor Shin}, \bibinfo{person}{Yasaman
  Razeghi}, \bibinfo{person}{Robert~L Logan~IV}, \bibinfo{person}{Eric
  Wallace}, {and} \bibinfo{person}{Sameer Singh}.}
  \bibinfo{year}{2020}\natexlab{}.
\newblock \showarticletitle{Autoprompt: Eliciting knowledge from language
  models with automatically generated prompts}.
\newblock \bibinfo{journal}{\emph{arXiv preprint arXiv:2010.15980}}
  (\bibinfo{year}{2020}).
\newblock


\bibitem[Sigurdsson et~al\mbox{.}(2018a)]%
        {sigurdsson2018actor}
\bibfield{author}{\bibinfo{person}{Gunnar~A Sigurdsson},
  \bibinfo{person}{Abhinav Gupta}, \bibinfo{person}{Cordelia Schmid},
  \bibinfo{person}{Ali Farhadi}, {and} \bibinfo{person}{Karteek Alahari}.}
  \bibinfo{year}{2018}\natexlab{a}.
\newblock \showarticletitle{Actor and observer: Joint modeling of first and
  third-person videos}. In \bibinfo{booktitle}{\emph{proceedings of the IEEE
  conference on computer vision and pattern recognition}}.
  \bibinfo{pages}{7396--7404}.
\newblock


\bibitem[Sigurdsson et~al\mbox{.}(2018b)]%
        {sigurdsson2018charades}
\bibfield{author}{\bibinfo{person}{Gunnar~A Sigurdsson},
  \bibinfo{person}{Abhinav Gupta}, \bibinfo{person}{Cordelia Schmid},
  \bibinfo{person}{Ali Farhadi}, {and} \bibinfo{person}{Karteek Alahari}.}
  \bibinfo{year}{2018}\natexlab{b}.
\newblock \showarticletitle{Charades-ego: A large-scale dataset of paired third
  and first person videos}.
\newblock \bibinfo{journal}{\emph{arXiv preprint arXiv:1804.09626}}
  (\bibinfo{year}{2018}).
\newblock


\bibitem[Sudhakaran et~al\mbox{.}(2019)]%
        {sudhakaran2019lsta}
\bibfield{author}{\bibinfo{person}{Swathikiran Sudhakaran},
  \bibinfo{person}{Sergio Escalera}, {and} \bibinfo{person}{Oswald Lanz}.}
  \bibinfo{year}{2019}\natexlab{}.
\newblock \showarticletitle{Lsta: Long short-term attention for egocentric
  action recognition}. In \bibinfo{booktitle}{\emph{Proceedings of the IEEE/CVF
  Conference on Computer Vision and Pattern Recognition}}.
  \bibinfo{pages}{9954--9963}.
\newblock


\bibitem[Sun et~al\mbox{.}(2020)]%
        {sun2020view}
\bibfield{author}{\bibinfo{person}{Jennifer~J Sun}, \bibinfo{person}{Jiaping
  Zhao}, \bibinfo{person}{Liang-Chieh Chen}, \bibinfo{person}{Florian Schroff},
  \bibinfo{person}{Hartwig Adam}, {and} \bibinfo{person}{Ting Liu}.}
  \bibinfo{year}{2020}\natexlab{}.
\newblock \showarticletitle{View-invariant probabilistic embedding for human
  pose}. In \bibinfo{booktitle}{\emph{Computer Vision--ECCV 2020: 16th European
  Conference, Glasgow, UK, August 23--28, 2020, Proceedings, Part V 16}}.
  Springer, \bibinfo{pages}{53--70}.
\newblock


\bibitem[Vyas et~al\mbox{.}(2020)]%
        {vyas2020multi}
\bibfield{author}{\bibinfo{person}{Shruti Vyas}, \bibinfo{person}{Yogesh~S
  Rawat}, {and} \bibinfo{person}{Mubarak Shah}.}
  \bibinfo{year}{2020}\natexlab{}.
\newblock \showarticletitle{Multi-view action recognition using cross-view
  video prediction}. In \bibinfo{booktitle}{\emph{Computer Vision--ECCV 2020:
  16th European Conference, Glasgow, UK, August 23--28, 2020, Proceedings, Part
  XXVII 16}}. Springer, \bibinfo{pages}{427--444}.
\newblock


\bibitem[Wang et~al\mbox{.}(2021)]%
        {wang2021tent}
\bibfield{author}{\bibinfo{person}{Dequan Wang}, \bibinfo{person}{Evan
  Shelhamer}, \bibinfo{person}{Shaoteng Liu}, \bibinfo{person}{Bruno
  Olshausen}, {and} \bibinfo{person}{Trevor Darrell}.}
  \bibinfo{year}{2021}\natexlab{}.
\newblock \showarticletitle{Tent: Fully Test-Time Adaptation by Entropy
  Minimization}. In \bibinfo{booktitle}{\emph{International Conference on
  Learning Representations}}.
\newblock
\urldef\tempurl%
\url{https://openreview.net/forum?id=uXl3bZLkr3c}
\showURL{%
\tempurl}


\bibitem[Wang et~al\mbox{.}(2016)]%
        {wang2016temporal}
\bibfield{author}{\bibinfo{person}{Limin Wang}, \bibinfo{person}{Yuanjun
  Xiong}, \bibinfo{person}{Zhe Wang}, \bibinfo{person}{Yu Qiao},
  \bibinfo{person}{Dahua Lin}, \bibinfo{person}{Xiaoou Tang}, {and}
  \bibinfo{person}{Luc Van~Gool}.} \bibinfo{year}{2016}\natexlab{}.
\newblock \showarticletitle{Temporal segment networks: Towards good practices
  for deep action recognition}. In \bibinfo{booktitle}{\emph{European
  conference on computer vision}}. Springer, \bibinfo{pages}{20--36}.
\newblock


\bibitem[Yang et~al\mbox{.}(2022b)]%
        {yang2022prompting}
\bibfield{author}{\bibinfo{person}{Jinyu Yang}, \bibinfo{person}{Zhe Li},
  \bibinfo{person}{Feng Zheng}, \bibinfo{person}{Ales Leonardis}, {and}
  \bibinfo{person}{Jingkuan Song}.} \bibinfo{year}{2022}\natexlab{b}.
\newblock \showarticletitle{Prompting for Multi-Modal Tracking}. In
  \bibinfo{booktitle}{\emph{Proceedings of the 30th ACM International
  Conference on Multimedia}}. \bibinfo{pages}{3492--3500}.
\newblock


\bibitem[Yang et~al\mbox{.}(2022a)]%
        {yang2022oakink}
\bibfield{author}{\bibinfo{person}{Lixin Yang}, \bibinfo{person}{Kailin Li},
  \bibinfo{person}{Xinyu Zhan}, \bibinfo{person}{Fei Wu},
  \bibinfo{person}{Anran Xu}, \bibinfo{person}{Liu Liu}, {and}
  \bibinfo{person}{Cewu Lu}.} \bibinfo{year}{2022}\natexlab{a}.
\newblock \showarticletitle{OakInk: A Large-scale Knowledge Repository for
  Understanding Hand-Object Interaction}. In
  \bibinfo{booktitle}{\emph{Proceedings of the IEEE/CVF Conference on Computer
  Vision and Pattern Recognition}}. \bibinfo{pages}{20953--20962}.
\newblock


\bibitem[Yang et~al\mbox{.}(2022c)]%
        {yang2022attracting}
\bibfield{author}{\bibinfo{person}{Shiqi Yang}, \bibinfo{person}{Yaxing Wang},
  \bibinfo{person}{Kai Wang}, \bibinfo{person}{Shangling Jui}, {et~al\mbox{.}}}
  \bibinfo{year}{2022}\natexlab{c}.
\newblock \showarticletitle{Attracting and dispersing: A simple approach for
  source-free domain adaptation}. In \bibinfo{booktitle}{\emph{Advances in
  Neural Information Processing Systems}}.
\newblock


\bibitem[Zhang et~al\mbox{.}(2022b)]%
        {zhang2022can}
\bibfield{author}{\bibinfo{person}{Renrui Zhang}, \bibinfo{person}{Ziyao Zeng},
  \bibinfo{person}{Ziyu Guo}, {and} \bibinfo{person}{Yafeng Li}.}
  \bibinfo{year}{2022}\natexlab{b}.
\newblock \showarticletitle{Can Language Understand Depth?}. In
  \bibinfo{booktitle}{\emph{Proceedings of the 30th ACM International
  Conference on Multimedia}}. \bibinfo{pages}{6868--6874}.
\newblock


\bibitem[Zhang et~al\mbox{.}(2022a)]%
        {zhang2022audio}
\bibfield{author}{\bibinfo{person}{Yunhua Zhang}, \bibinfo{person}{Hazel
  Doughty}, \bibinfo{person}{Ling Shao}, {and} \bibinfo{person}{Cees~GM
  Snoek}.} \bibinfo{year}{2022}\natexlab{a}.
\newblock \showarticletitle{Audio-adaptive activity recognition across video
  domains}. In \bibinfo{booktitle}{\emph{Proceedings of the IEEE/CVF Conference
  on Computer Vision and Pattern Recognition}}. \bibinfo{pages}{13791--13800}.
\newblock


\bibitem[Zheng et~al\mbox{.}(2022)]%
        {zheng2022prompt}
\bibfield{author}{\bibinfo{person}{Zangwei Zheng}, \bibinfo{person}{Xiangyu
  Yue}, \bibinfo{person}{Kai Wang}, {and} \bibinfo{person}{Yang You}.}
  \bibinfo{year}{2022}\natexlab{}.
\newblock \showarticletitle{Prompt vision transformer for domain
  generalization}.
\newblock \bibinfo{journal}{\emph{arXiv preprint arXiv:2208.08914}}
  (\bibinfo{year}{2022}).
\newblock


\bibitem[Zhu et~al\mbox{.}(2023)]%
        {zhu2023visual}
\bibfield{author}{\bibinfo{person}{Jiawen Zhu}, \bibinfo{person}{Simiao Lai},
  \bibinfo{person}{Xin Chen}, \bibinfo{person}{Dong Wang}, {and}
  \bibinfo{person}{Huchuan Lu}.} \bibinfo{year}{2023}\natexlab{}.
\newblock \bibinfo{title}{Visual Prompt Multi-Modal Tracking}.
\newblock
\newblock
\showeprint[arxiv]{2303.10826}~[cs.CV]


\end{thebibliography}

\appendix

\newpage

\begin{figure*}
\centering
 \includegraphics[width=0.9\textwidth]{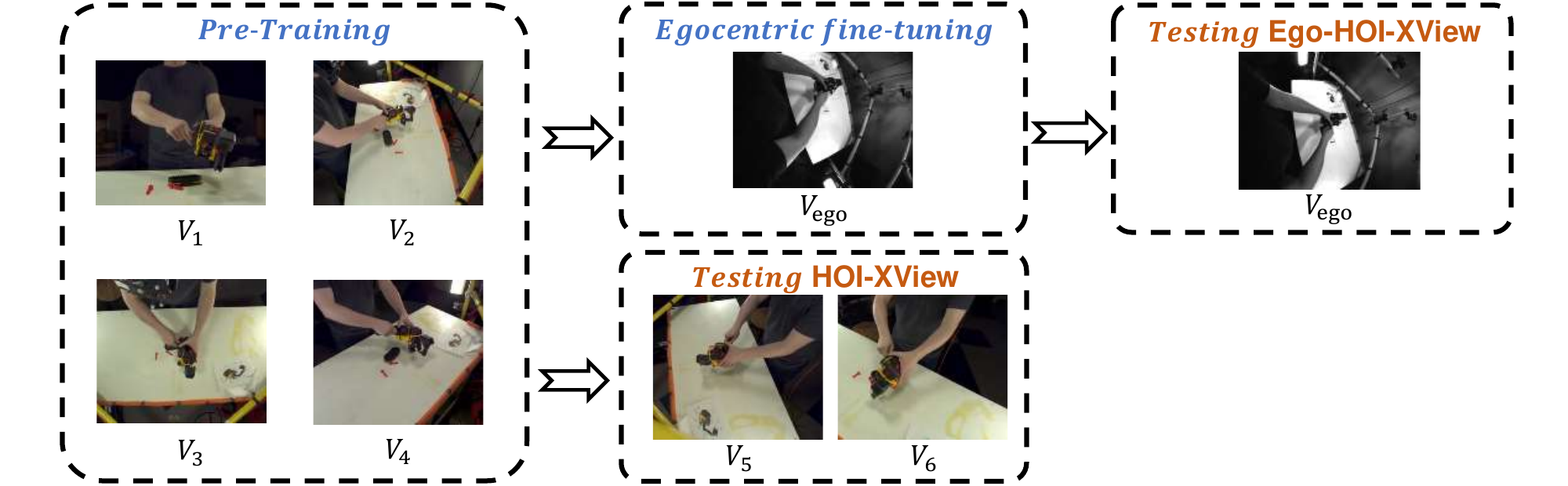}
\caption{Illustration of the training and inference pipeline of our POV on Assembly101 and its view split.
}
	\label{fig:assembly101}
\end{figure*}

\begin{figure*}
\centering
 \includegraphics[width=0.9\textwidth]{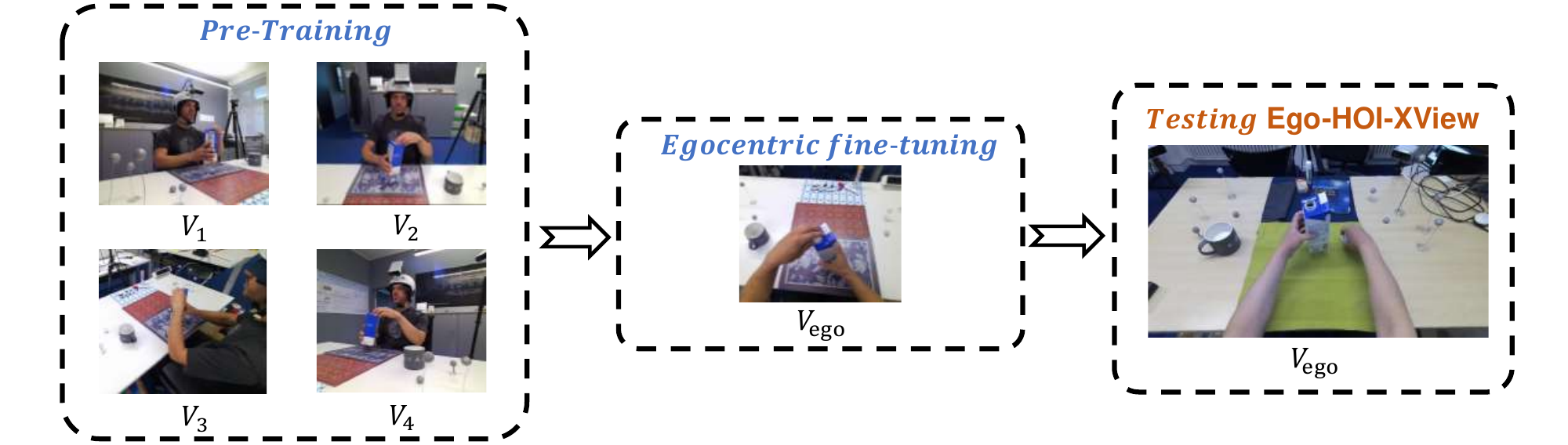}
\caption{Illustration of the training and inference pipeline of our POV on H2O and its view split.
}
	\label{fig:h2o}
\end{figure*}

\section{Additional Dataset Details}
\label{sec:dataset}

\noindent\textbf{Camera View Split. }
In Figure~\ref{fig:assembly101} and Figure~\ref{fig:h2o}, we show the entire training and inference pipeline of our method. 
Specifically, for both Assembly and H2O datasets, we use videos from four third-person views ($V_1\sim V_4$) for pre-training, then use $V_{\rm ego}$ for ego-centric fine-tuning and Ego-HOI-XView evaluation.
On Assembly101, we additionally utilize two extra third-person views $(V_5\sim V_6)$ for the HOI-XView evaluation, following the cross-view setup proposed by Das~\etal~\cite{das2019toyota}.
Note that although the camera views are synchronized and fixed, we are not using the synchronization information.

\noindent\textbf{Selected Pre-training Categories. }
The Assembly101 dataset with a total of 1380 classes has a severe long-tailed distribution, meaning that there are a small number of categories with many examples and a large number of categories with only a few examples.
Since the focus of this work is not to address the long-tailed distribution problem, we select a subset of 142 fine-grained head categories from the 1380 classes.
These fine-grained head categories are formed by combining 20 verbs and 24 nouns, as shown in Table~\ref{tab:class_assembly101}.
In contrast, the H2O dataset has a uniform category distribution, meaning that there are roughly an equal number of training samples for each category.
The fine-grained action categories in the H2O dataset are formed by combining 11 verbs and 8 nouns, as illustrated in Table~\ref{tab:h2o}.

\begin{figure*}[ht]
  \centering
  \includegraphics[width=\textwidth]{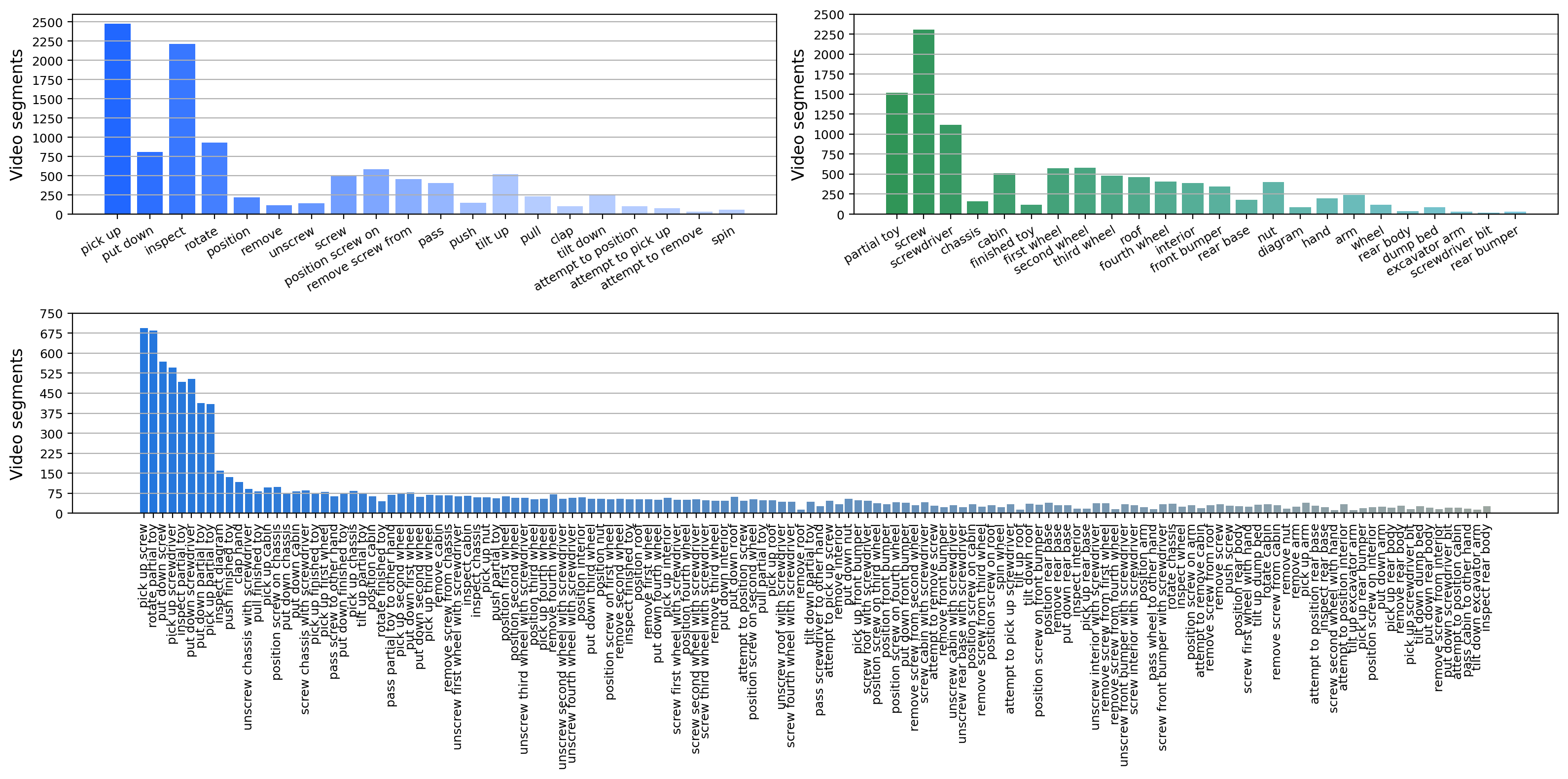}
  \caption{Distribution of verbs (top left), nouns (top right), and actions (bottom) on Assembly101.}
  \vspace{5pt}
  \label{fig:twosubs}
\end{figure*}

\begin{figure*}[ht]
  \centering
  \includegraphics[width=\textwidth]{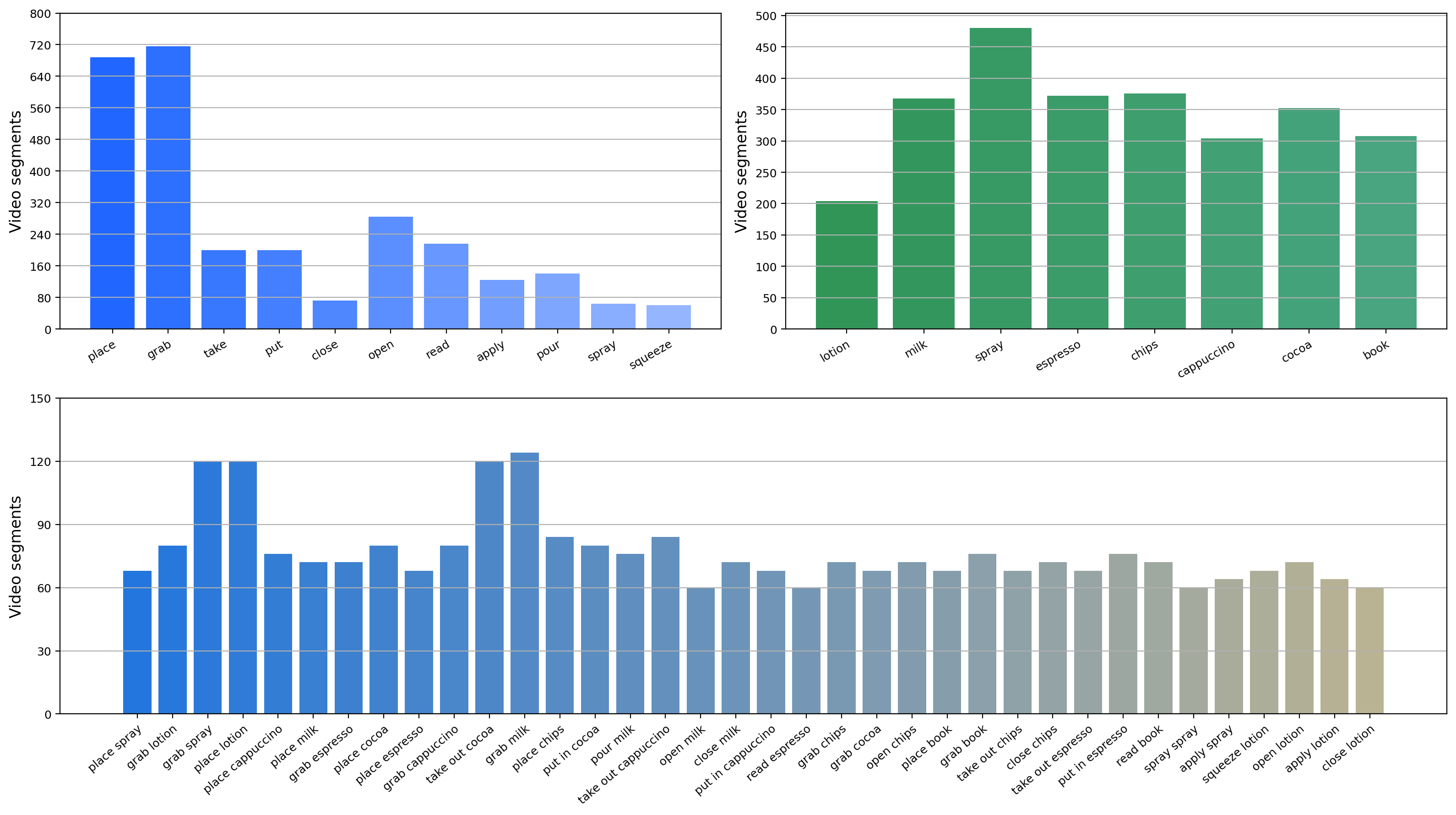}
  \caption{Distribution of verbs (top left), nouns (top right), and actions (bottom) on Assembly101.}
  \vspace{5pt}
  \label{fig:h2o_subs}
\end{figure*}

\noindent\textbf{Category Distribution. }
Figure~\ref{fig:twosubs} illustrates the category distribution of verbs, nouns, and actions for pre-training on Assembly101.
To perform additional fine-tuning, a subset of egocentric videos with an average of 73 samples per action class is selected.
For H2O, the distribution of video segments in each category is almost uniform, as illustrated in Figure~\ref{fig:h2o_subs}. 
For further egocentric fine-tuning, we select a subset of egocentric videos with an average of 5 samples per action class.

\section{Additional Experimental Details}
\label{sec:implement}

In this section, more experiment details are presented for the setups of (1) zero-shot Ego-HOI-XView recognition; (2) few-shot Ego-HOI-XView recognition, and (3) 3rd-to-ego HOI recognition.


\subsection{Zero-shot Ego-HOI-XView Recognition}
The zero-shot Ego-HOI-XView recognition task in this paper can be regarded as a domain adaption problem, which 
generally uses unlabeled target domain data to adapt a model trained on source data to new domains.
In the experiments of zero-shot Ego-HOI-XView Recognition, we re-implement two types of domain-adaption approaches for comparison: (1) fine-tuning most parameters (e.g., AaD~\cite{yang2022attracting} and SHOT~\cite{liang2020we}), and (2) fine-tuning only a small number of parameters (e.g., Tent~\cite{wang2021tent}).
Here, we introduce the details of their implementation which are omitted in the main paper due to space limitations:

\noindent\textbf{(1) SHOT~\cite{liang2020we}}, which fine-tunes on unlabeled target data to address domain adaption by minimizing information maximization loss and deep clustering.
SHOT fine-tunes the entire model except for the classification head. 
We opt not to utilize the suggested weight normalization, batch normalization, or label smoothing techniques in our implementation, as they are irrelevant to the scope of our study.

\noindent\textbf{(2) AaD~\cite{yang2022attracting}}, which is a state-of-the-art method for domain adaption that optimizes an upper bound of its clustering objective. It fine-tunes the full parameters as well.

\noindent\textbf{(3) Tent~\cite{wang2021tent}}, which adopts information entropy loss to simply fine-tune normalization layer parameters in CNNs.
Specifically, we adjust transformation parameters in Layer Normalization used in vision transformers instead of Batch Normalization used in CNNs.

In all of the aforementioned methods, we employ an AdamW optimizer with an initial learning rate of 1e-4 during training. We utilize a cosine scheduler whose learning rate gradually descends to 1e-6 as the final learning rate.

\begin{table}[h]
\caption{Categories of verbs and nouns in Assembly101.}
\begin{tabular}{c|c| c@{} c|c}
\toprule
ID & verb&& ID & noun \\
\midrule
0&pick up&&0 & screw              \\
1&rotate&&1 & partial toy        \\
2&put down&&2 & screwdriver        \\
3&inspect&&3 & diagram            \\
4&push&&4 & finished toy       \\
5&clap&&5 & hand               \\
6&pull&&6 & cabin              \\
7&unscrew&&7 & chassis           \\
8&position&&8 & first wheel        \\
9&screw&&9 & second wheel       \\
10&position screw on&&10 &  third wheel      \\
11&tilt up&&11 &  fourth wheel     \\
12&remove&&12 &  interior          \\
13&pass&&13 & nut               \\
14&tilt down&&14 &roof              \\
15&remove screw from&&15 &wheel            \\
16&attempt to position&&16 &rear base        \\
17&attempt to pick up&&17 &front bumper      \\
18&spin&&18 &arm               \\
19& attempt to remove&&19 &dump bed          \\
-&-&&20 &rear body         \\
-&-&&21 &excavator arm    \\
-&-&&22 &rear bumper       \\
-&-&&23 &screwdriver bit  \\
\bottomrule
\end{tabular}\label{tab:class_assembly101}
\end{table}

\begin{table}[]
\caption{Categories of verbs and nouns in H2O.}
\begin{tabular}{c|c| c@{} c|c}
\toprule
ID & verb&& ID & noun \\
\midrule
0&grab&&0 & book              \\
1&place&&1 & espresso        \\
2&open&&2 & lotion        \\
3&close&&3 & spray            \\
4&pour&&4 & milk       \\
5&take out&&5 & cocoa               \\
6&put in&&6 & chips              \\
7&apply&&7 & cappuccino           \\
8&read&&- & -       \\
9&spray&&- & -       \\
10&squeeze&&- &  -      \\
\bottomrule
\end{tabular}\label{tab:h2o}
\end{table}

\subsection{Few-shot Ego-HOI-XView Recognition}
Under the few-shot Ego-HOI-XView recognition setup, the model is fine-tuned on labeled egocentric videos using full parameters. 
We compare our POV with two pre-training methods that are fine-tuned on limited egocentric videos:

\noindent\textbf{(1) EgoVLP~\cite{lin2022egocentric}}, which is a CLIP-style model pre-trained on Ego4D~\cite{grauman2022ego4d} with over 3.8 million egocentric video-text pairs. We use the publicly released pre-trained model without further fine-tuning. Then, we use an AdamW optimizer with a learning rate of 2e-4 and fine-tune it for 20 epochs while leveraging early stopping for better results.

\begin{figure*}[ht]
  \centering
  \includegraphics[width=0.80\textwidth]{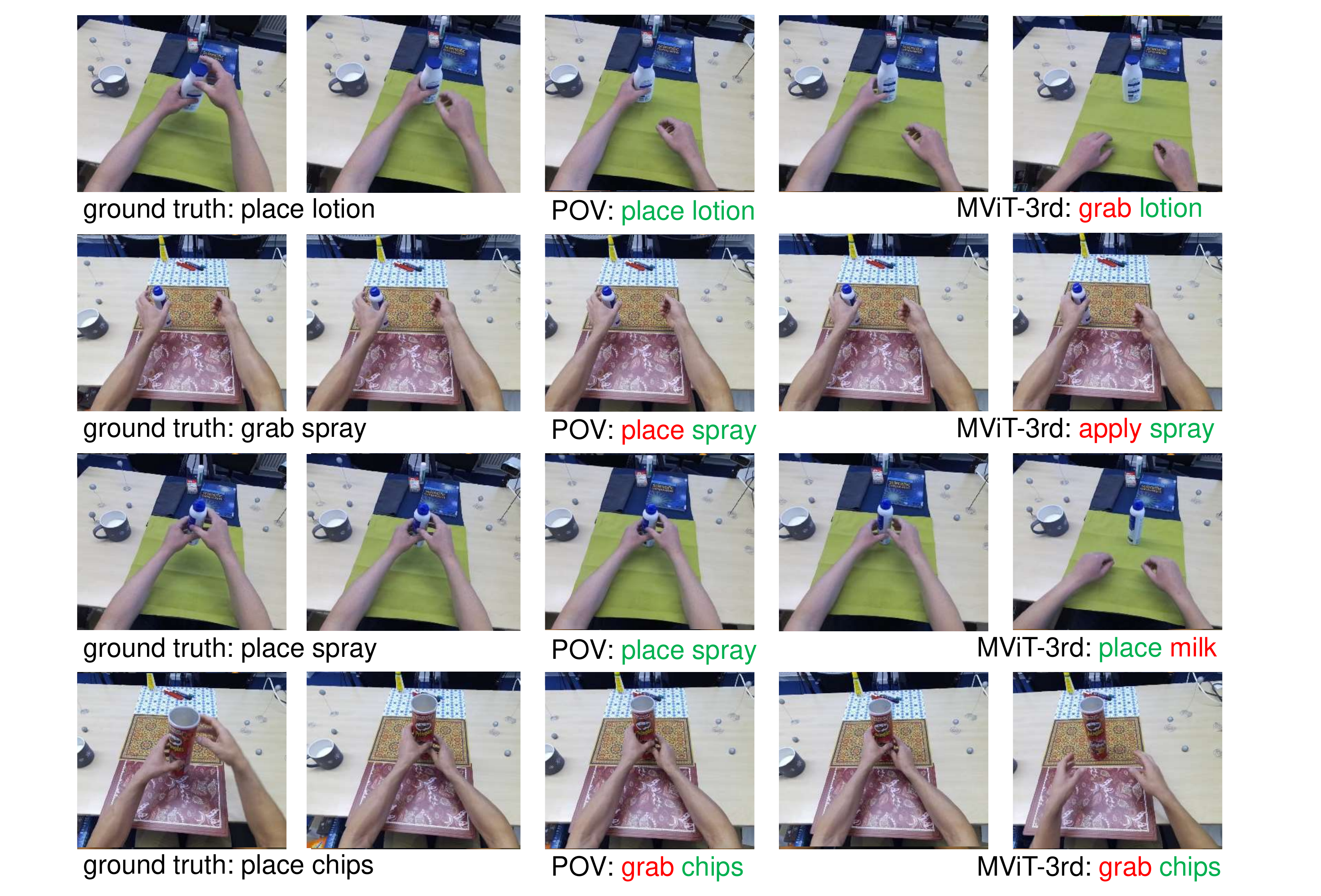}
  \caption{Some examples of egocentric predictions on H2O.}
  \vspace{-5pt}
  \label{fig:quanlitative_h2o}
\end{figure*}
\begin{figure*}[ht]
  \centering
  \includegraphics[width=0.8\textwidth]{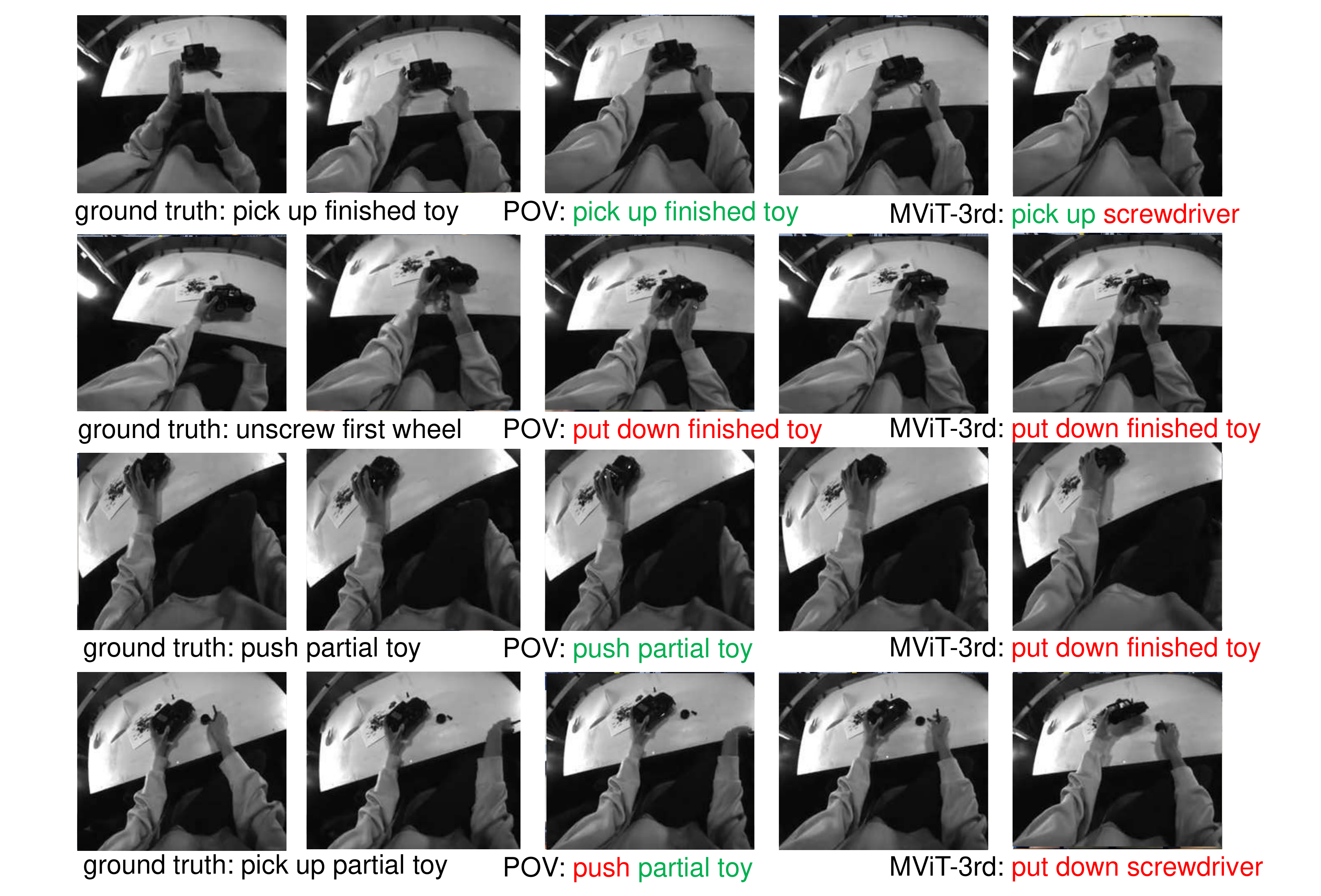}
  \caption{Some examples of egocentric predictions on Assembly101.}
  \vspace{-5pt}
  \label{fig:quanlitative_assembly101}
\end{figure*}
\noindent\textbf{(2) Ego-Exo~\cite{li2021ego}}, which is pre-trained on the large-scale video dataset Kinetics-400 with over 1 million third-person videos, exploring useful visual cues for egocentric action recognition~\cite{li2021ego}. We use the pre-trained SlowFast ResNet50 architecture, whose parameters are close to those of MViT-S that we use at $\sim$30M, for further fine-tuning on our limited egocentric videos. We use an SGD optimizer with a starting learning rate of 1e-2 and fine-tune it for 30 epochs with the learning rate dropping by 0.1 at the 20th and 25th epochs.

\subsection{3rd-to-ego HOI Recognition}
In our ablation study of egocentric fine-tuning, 3rd-to-ego HOI recognition requires the model to predict hand-object interactions in egocentric videos directly, without egocentric fine-tuning after pre-training on third-person videos. We implement one view-agnostic learning method and one zero-shot egocentric video learner.

\noindent\textbf{(1) 3DTRL~\cite{shang2022learning}}, which is an explicit geometric transformation layer that is inserted between transformer layers. We add the 3DTRL layer after the first transformer layer. The model is trained using the AdamW optimizer with a learning rate starting at 1e-3 and ending at 1e-5 using a cosine scheduler.

\noindent\textbf{(2) EgoVLP~\cite{lin2022egocentric}}, which has excellent zero-shot recognition ability. We follow the zero-shot setting in this task suggested by Lin~\etal~\cite{lin2022egocentric}, where video classification is performed by calculating the video-text similarities and matching video to the most similar text of action class.

\section{Quanlitative results}
\label{sec:quanlitative}
We present additional qualitative results for Ego-HOI-XView on H2O and Assembly101 in this section.
As illustrated in Figure~\ref{fig:quanlitative_h2o} and Figure~\ref{fig:quanlitative_assembly101}, our POV enhances the accuracy of recognizing verbs and nouns.
However, distinguishing between certain verbs in daily life, such as "grab" and "place", poses a challenge, whose prediction results strongly relies on temporal information.
Furthermore, on specific domains in Assembly101, there still remains a tough problem on recognizing hand-object interaction.









\end{document}